\documentclass[review]{elsarticle}

\usepackage{floatrow}
\newfloatcommand{capbtabbox}{table}[][\FBwidth]

\usepackage{blindtext}

\usepackage{hyperref}
\usepackage{mathtools}

\usepackage{color}
\usepackage{tcolorbox}
\usepackage{amsmath}
\usepackage{amsfonts}
\usepackage{dsfont}
\usepackage{amssymb}
\usepackage{graphicx}
\usepackage{enumitem}
\usepackage{float}
\usepackage{longtable}
\usepackage{algcompatible}
\usepackage{algorithm}
\usepackage{caption}
\usepackage{adjustbox}
\usepackage{array}
\usepackage{rotating} 
\usepackage[font=scriptsize, skip=0cm, labelfont=bf]{caption}
\usepackage{booktabs}
\usepackage[justification=centering]{caption}
\usepackage[title]{appendix}
\usepackage{blindtext}

\usepackage{geometry}

\usepackage[utf8]{inputenc}
\usepackage[english]{babel}

\newcommand{\proofofref}{}
\newproof{zproofof}{Proof of \proofofref}

\usepackage{tikz}

\usepackage{xurl}
\usepackage{caption}
\usepackage{subcaption}

\journal{arXiv.org}

\biboptions{sort&compress}
\begin{document}

\begin{frontmatter}

\title{A machine learning analysis of the relationship between some underlying medical conditions and COVID-19 susceptibility}

\author[label1]{Mostafa Rezapour}
\ead{rezapom@wfu.edu}
\author[label1]{Colin A. Varady}
\ead{varaca19@wfu.edu}

\address[label1]{Department of Mathematics and Statistics, Wake Forest University, NC, U.S.}

\begin{abstract}
For the past couple years, the Coronavirus, commonly known as COVID-19, has significantly affected the daily lives of all citizens residing in the United States by imposing several, fatal health risks that cannot go unnoticed. In response to the growing fear and danger COVID-19 inflicts upon societies in the USA, several vaccines and boosters have been created as a permanent remedy for individuals to take advantage of. In this paper, we investigate the relationship between the COVID-19 vaccines and boosters and the total case count for the Coronavirus across multiple states in the USA. Additionally, this paper discusses the relationship between several, selected underlying health conditions with COVID-19. To discuss these relationships effectively, this paper will utilize statistical tests and machine learning methods for analysis and discussion purposes. Furthermore, this paper reflects upon conclusions made about the relationship between educational attainment, race, and COVID-19 and the possible connections that can be established with underlying health conditions, vaccination rates, and COVID-19 total case and death counts.
\end{abstract}


\end{frontmatter}

\section{Introduction}
The Coronavirus (COVID-19), caused by the SARS-CoV-2 virus, is an infectious disease that has changed the course of our history primarily due to varying health risks and symptoms associated with contracting the disease. The United States and other countries across the world have taken several strict, precautionary measures to slow the transmission of the virus by imposing mask mandates, social distance guidelines, and other sanitization recommendations. Moreover, several organizations such as the CDC and HRSA have advocated for two-dose vaccines and single-dose boosters to mitigate the cases and deaths associated with the Coronavirus disease. Some of these two-dose vaccines promoted by health organizations include the Pfizer, Moderna, and Johnson and Johnson (J\&J) vaccines. However, with the abundance of protective measures individuals can now take to immunize themselves from COVID-19, will these vaccines fully and efficiently mitigate the health risks and symptoms associated with this infectious disease?

Over the past couple months, new highly contagious and dangerous COVID-19 variants have begun to arise, specifically the Omicron and Delta variants. As a result, there have been numerous reports of breakthrough cases in countries such as the United States. According to the Center of Disease Control (CDC), a breakthrough case is defined as, “someone who has detectable levels of SARS-CoV-2, the novel coronavirus, in their body at least 14 days after they’ve been fully vaccinated against COVID-19.” Therefore, with the growing numbers of breakthrough cases arising across the United States, health companies such as the World Health Organization (WHO) have recommended that individuals obtain an additional “booster” shot of the COVID-19 vaccine. Even with the additional booster shots now being recommended by these organizations, will individuals still be sufficiently protected against the dangers associated with the Coronavirus?

Since early 2020, after a December 2019 outbreak in China, Machine learning (ML) has been employed in the fight against the Coronavirus disease. Machine Learning methods are categorized into two areas: Unsupervised Learning and Supervised Learning methods. Unsupervised learning is a type of machine learning in which an algorithm is utilized to analyze and cluster unlabeled datasets. Unsupervised machine learning and statistical learning algorithms can be used to find out the relationship between variables in a dataset.  On the other hand, supervised learning algorithms can learn, and generalize from historical data in order to make predictions on new data. Additionally, supervised learning machine learning algorithms can also be used to identify the most related features to the target variable. Optimization methods (e.g. \cite{1,2,3,4,5,6,7,8,9,10}), which are often used to minimize a loss or error function in the model training process, play a significant role in the speed-accuracy trade-off of machine learning algorithms.  

Kushwaha et al. \cite{11} reviews the relevant papers on machine learning for COVID-19 from the databases of SCOPUS, Academia, Google Scholar, PubMed, and ResearchGate. To evaluate the research trends in the coronavirus disease (COVID-19), De Felice et al. \cite{12} performs bibliometric analysis using a machine learning bibliometric methodology. Alimadadi et al. \cite{13} reviews AI and ML methods that have been employed in the fight against COVID-19. Elaziz et al. \cite {14} proposes a new machine learning method to classify the chest x-ray images into two classes: COVID-19 patient or non-COVID-19 person. Punn et al. \cite{15} utilizes machine learning and deep learning models to have a better understanding of exponential behavior along with the prediction of future reachability of the COVID-19. Sujath et al. \cite{16} performs linear regression, Multilayer perceptron and Vector autoregression methods to analyze the COVID-19 pandemic in India.  Lalmuanawma et al \cite{17} comprehensively reviews multiple AI and ML methods that have been employed for screening, predicting, forecasting, contact tracing, and drug development for SARS-CoV-2. Cheng et al \cite{18} uses time series data, including vital signs, nursing assessments, laboratory data, and electrocardiograms to train a random forest (RF) model to be used as a screening tool to identify patients at risk of imminent ICU transfer within 24 hours. Rezapour et al. \cite{19} utilizes Decision Trees, Multinomial Logistic Regression, Naive Bayes, k-Nearest Neighbors, Support Vector Machines, Neural Networks, Random Forests, Gradient Tree Boosting, XGBoost, CatBoost, LightGBM, Synthetic Minority Oversampling, and Chi-Squared tests to analyze the impacts the COVID-19 pandemic has had on the mental health of frontline workers in the United States. Rezapour \cite{20} employs multiple supervised and unsupervised methods to find out relationships between COVID-19 related negative effects and alcohol use changes in healthcare workers. 

In this paper, we utilize Gradient Boosting Decision Trees (GBDT) based methods to find out the relationship between several, selected underlying health conditions with COVID-19. By means of GBDT-based models, the first question this paper attempts to answer is whether the COVID-19 vaccine doses or the COVID-19 boosters provide sufficient protection against the array of COVID-19 variants. Statistically speaking, is there a strong, negative relationship between the COVID-19 vaccination rates and the COVID-19 confirmed cases? The CDC has completed countless studies on the effectiveness of the Pfizer, and Moderna vaccines as well as the efficacy of obtaining an additional “booster” vaccine. Concerning the effectiveness of the Pfizer and Moderna vaccines, according to the MMWR Early Release posted by the CDC,  the two-dose COVID-19 vaccines provide sufficient protection against COVID-19 hospitalizations, specifically a protection rate of 93\% for Moderna and 88\% for Pfizer \cite{CDC 1}. According to their findings, all FDA-approved or authorized COVID-19 vaccines provide substantial protection against COVID-19 hospitalization while there could be some variation in effectiveness of the protective measures these vaccines promise to produce for the patient \cite{CDC 1}. However, the study does not address whether these vaccines are still effective in patients with “immunocompromising conditions” \cite{CDC 1}. In this paper, we utilize multiple supervised and unsupervised machine learning methods to address the relationship between underlying health conditions, such as diabetes, heart disease, and septicemia, and the total death count due to COVID-19. 

By means of GBDT-based models, the second question that this paper attempts to answer is whether an underlying health condition contributes to the fatality rate of COVID-19 across the United States. Statistically speaking, is there a strong, positive relationship between having an underlying health condition and COVID-19 related deaths? In this paper, we also utilize machine learning methods to find out whether having an underlying health condition enhances the symptoms associated with COVID-19. In conjunction to the analysis of the relationship between underlying health conditions and the total death count due to COVID-19, this paper also seeks to establish a relationship between educational attainment and COVID-19 related deaths. More specifically, this paper seeks to identify a connection between grade level, ethnicity, and COVID-19 deaths. 

The remainder of this paper is structured as follows: In the following section, a review of GBDT methods that we utilize in this paper is provided. Section \ref{section3} discusses the methodology, describes the experimental framework used to find out the relationship between several, selected underlying health conditions with COVID-19. Section \ref{section4} discusses the results obtained by Machine Learning methods. Finally, Section \ref{section5} concludes the paper by summarizing our overall findings. 

\section{Gradient Boosting Decision Trees Review}

According to Kaggle, “the most commonly used algorithms were linear and logistic regression, followed closely by decision trees and random forests. Of more complex methods, gradient boosting machines and convolutional neural networks were the most popular approaches” \cite{11-n}. Gradient boosting machines, or simply, GBMs \cite {11-n1}, are powerful machine-learning techniques that combine the predictions from multiple weak learners, which are decision trees, to predict outcomes correctly. Unlike Bagging \cite{breiman1996bagging}, like Random Forests \cite{Breiman}, in which trees are constructed to their maximum extent, Gradient Boosting Machine constructs small decision trees, which are not very deep, sequentially such that each subsequent tree aims to reduce the errors of the previous tree \cite {11-n1}. Gradient Boosting Decision Tree (GBDT) \cite {friedman2001greedy} is a GBM method in which Decision Trees (weak learners) are ensembled and trained in sequence. One of the most important factors in GBDT’s speed is how to find the best split points. There are several algorithms for finding the best split points such as the pre-sorted algorithm \cite{GBDTTREE1, GBDTTREE2} and the histogram-based algorithm \cite{GBDTTREE3, GBDTTREE4, GBDTTREE5}. The most popular boosting algorithms that follow GBMs’ framework (or GBDT's framework) are XGBoost (introduced by Chen et al. \cite{34,35}), LightGBM (introduced by Ke et al. \cite{36}), and Catboost (introduced by Dorogush, et al. \cite{37,37-1}). 

XGBoost is a scalable machine learning system for tree boosting and one of the most popular decision-tree-based ensemble Machine Learning algorithms \cite{35}. It is used widely by data scientists, e.g. in Kaggle competitions and DataHack hackathons, due to its scalability in all scenarios, high execution speed and very good performance. The purpose of this subsection is to provide an overview of XGBoost. 

Suppose a training data set, namely $D$, with $n$ instances, and $m$ features, 
$
D=\big\{(x_i,y_i) \big | \text{ }x_i \in \mathbb{R}^m, \text{ }y_i \in \mathbb{R}\big\}
$
is given, and we aim to train a XGBoost with $K$ weak learners, decision trees $q_1, q_2, \dots, q_K$ with independent tree structures $f_1, f_2, \dots, f_K$, to predict the output (the target variable $y$). To train the model, we need to minimize the regularized overall loss function,
\begin{equation}\label{eq1}
\mathcal{L}=\sum_{i=1}^{n}\ell(y_i,\hat{y}_i)+\sum_{j=1}^{K}\Omega(f_j),
\end{equation}
where $\ell$ is a differentiable convex loss function that measures the difference between the prediction $y_i$ and the target $\hat{y}_i$, $\Omega(f_j)= \gamma T_j+\frac{1}{2}\lambda \|\omega_j\|^2$, $T_j$ is the number of leaves in the
tree $f_j$, $\omega_j=(\omega_j^1, \omega_j^2, \dots, \omega_j^{T_j}) \in \mathbb{R}^{T_j}$ is leaf weights vectors, and $\omega_j^p \in \mathbb{R}$, for all $p= 1, 2, \dots, T_j$, represents weight score on p-th leaf, $\gamma$ and $\lambda$ are regularization parameters, and $\hat{y}_i=\phi(x_i)=\sum_{j=1}^{K} f_j(x_i)$. Note that the regularization term $\sum_{j=1}^{K}\Omega(f_j)$, which helps the model avoid overfitting, differs XGBoost from other gradient boosting methods \cite{34}. 

\begin{figure}[H]
  \includegraphics[width=.8\linewidth]{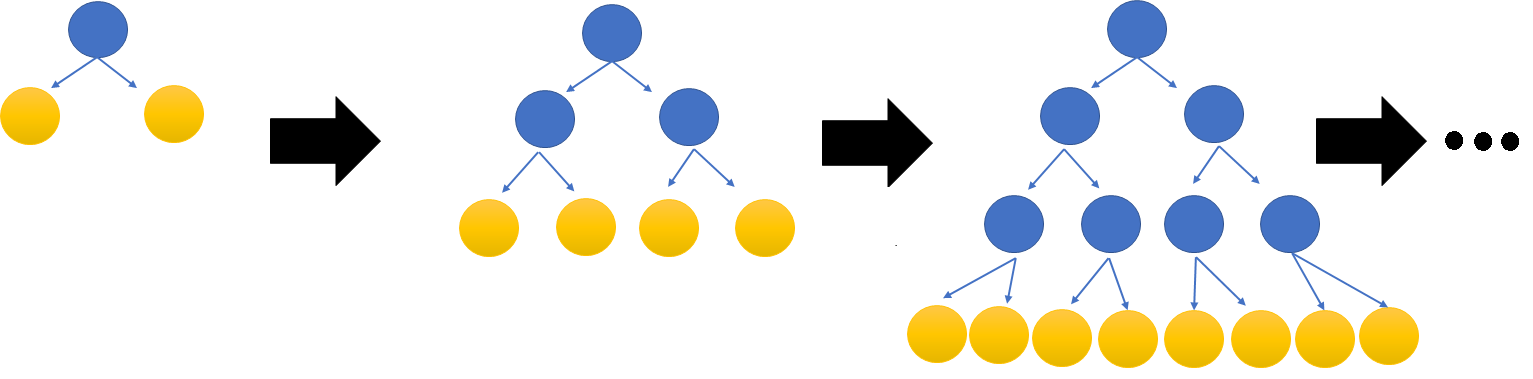}
  \caption{XGBoost uses the layer-wise tree growth strategy}
  \label{fig:XG1}
\end{figure}

We look for the optimal tree structures $f_1, f_2, \dots, f_K$ such that $\mathcal{L}$,  in Equation (\ref{eq1}), meets its minimum. But, since the objective function $\mathcal{L}$ includes functions as parameters, we cannot minimize it using the traditional optimization methods. Instead, Chen et al. \cite{34} minimizes the objective function $\mathcal{L}$ in an additive manner. Let $\hat{y}_i^{(t)}$ be the
prediction of the i-th instance at the t-th iteration, and $\hat{y}_i^{(0)}=0$. Hence, Equation (\ref{eq1}) can be restated as 
\begin{equation}\label{eq2}
\mathcal{L}=\sum_{t=1}^{K}\mathcal{L}^{(t)},
\end{equation}
where,
\begin{equation}\label{eq3}
\mathcal{L}^{(t)}=\sum_{i=1}^{n}\ell(y_i,\hat{y}_i^{(t-1)}+f_t(x_i))+\Omega(f_t).
\end{equation}
Note that  $\ell(y_i,\hat{y}_i^{(t-1)}+f_t(x_i))$ is smooth, and can be approximated and replaced with the quadratic model 
\begin{equation}\label{eq4}
\ell(y_i,\hat{y}_i^{(t-1)}+f_t(x_i))\approx \tilde{\ell}(y_i,\hat{y}_i^{(t-1)}+f_t(x_i))=\ell(y_i,\hat{y}_i^{(t-1)})+g_if_t(x_i)+\frac{1}{2}H_if_t(x_i)^2,
\end{equation}
where $g_i=\partial_{\hat{y}_i^{(t-1)}}\ell(y_i,\hat{y}_i^{(t-1)})$ and $H_i=\partial^2_{\hat{y}_i^{(t-1)}}\ell(y_i,\hat{y}_i^{(t-1)})$ are first and second order gradient statistics on the loss function \cite{34}. Therefore, the loss function $\mathcal{L}^{(t)}$ in Equation (\ref{eq3}) can be approximated by 
\begin{equation}\label{eq5}
\tilde{\mathcal{L}}^{(t)}=\sum_{i=1}^{n}\tilde{\ell}(y_i,\hat{y}_i^{(t-1)}+f_t(x_i))+\Omega(f_t).
\end{equation}
For a given instance set of leaf $I_j=\{i | \text{} q(x_i)=j\}$, it turns out the optimal weight $\bar{\omega}_t^j$ of leaf $j$ is
\begin{equation}\label{eq6}
\bar{\omega}_t^j=-\frac{\sum_{i \in I_j}g_i}{\sum_{i \in I_j}H_i+\lambda},
\end{equation}
that results in the new gain function 
\begin{equation}\label{eq7}
G=\frac{1}{2}\bigg[\frac{\big(\sum_{i \in I_L}g_i\big)^2}{\sum_{i \in I_L}H_i+\lambda}+\frac{\big(\sum_{i \in I_R}g_i\big)^2}{\sum_{i \in I_R}H_i+\lambda}-\frac{\big(\sum_{i \in I}g_i\big)^2}{\sum_{i \in I}H_i+\lambda}\bigg]-\gamma,
\end{equation}
where $I_L$, $I_R$ are the instance sets of left and right nodes after the split, and $I=I_L \cup I_R$. Figure \ref{fig:XG1} displays how XGBoost constructs trees sequentially layer-wise.
Note that XGBoost supports both the pre-sorted algorithm, which enumerates all possible split points on the pre-sorted
feature values \cite{GBDTTREE1, GBDTTREE2}, and the histogram-based algorithm, which finds the best split points based on the feature histograms \cite{GBDTTREE3, GBDTTREE4, GBDTTREE5}. It is known that for non-sparse large-scale datasets, the histogram-based algorithm is more efficient than the pre-sorted algorithm, but, it is still very time-consuming because it costs $\mathcal{O}(nm)$ for histogram building, and $\mathcal{O}(bm)$ for split point finding, where $n$ is the number of instances, $m$ is the number of feature, and $b$ is the number of bins \cite{36}.

LightGBM, short for Light Gradient Boosting Machine, is Gradient Boosting Decision Tree (GBDT) \cite {friedman2001greedy} that utilizes two techniques, Gradient-based One-Side Sampling (GOSS) and Exclusive Feature Bundling (EFB), to reduce the number of data instances and the number of features. Since GBDT computational complexity is proportional to both the number of features and the number of instances, GOSS and EFB help LightGBM handle big data more efficiently. 

GOSS aims to achieve a good balance between reducing the number of data instances and keeping the accuracy for learned decision trees by keeping all the instances with large gradients and removing instances with small gradients. Instances with small gradients are those that are already well-trained, in other words, the training error for them is small. To earn more information from the under-trained instances, GOSS sorts the data instances according to the absolute value of their gradients and selects the top $a \times 100\%$ instances, namely $A$, then it randomly samples a subset $B$ containing $b\times 100\%$ instances from the rest of the data, which means the size of $B$ is $b\times |A^C|$. After that, GOSS amplifies the sampled data with small gradients by a constant $\frac{1-a}{b}$ in calculating the information gain \cite{36}. Suppose a training set $D$ contains $n$ instances, $x_1, x_2, \dots, x_n$, and let $g_1, g_2, \dots, g_n$ be the negative gradients of the loss function with respect to the output. The instances are split according to the estimated variance gain for feature $j$,
\begin{equation}
\tilde{V}_j(d)=\frac{1}{n}\Bigg(\frac{\big(\sum_{x_i \in A_L}g_i+\frac{1-a}{b}\sum_{x_i \in B_L}g_i\big)^2}{n_L^j(d)}+\frac{\big(\sum_{x_i \in A_R}g_i+\frac{1-a}{b}\sum_{x_i \in B_R}g_i\big)^2}{n_R^j(d)}\Bigg),
\end{equation}
where $A_L=\{x_i \in A : \text{} x_{ij} \leq d\}$, $A_R=\{x_i \in A : \text{} x_{ij} >d\}$, $B_L=\{x_i \in B : \text{} x_{ij} \leq d\}$, $B_R=\{x_i \in B : \text{} x_{ij} >d\}$, $n_L^j(d)=\sum \mathds{1}\big(x_i \in A \cup B : \text{} x_{ij} \leq d\big)$ and $n_R^j(d)=\sum \mathds{1}\big(x_i \in A \cup B : \text{} x_{ij} >d\big)$ with
\begin{equation}
\mathds{1} (P)=\begin{cases} 
          1& \text{if P is true,} \\
          0& \text{otherwise.}
       \end{cases}
\end{equation}
Note that $x_i=(x_{i1}, x_{i2}, \dots, x_{im})$ is a vector of $m$ numerical features, and $x_{ij}$ is the j-th component of $x_i$.

\begin{figure}[H]
  \includegraphics[width=.9\linewidth]{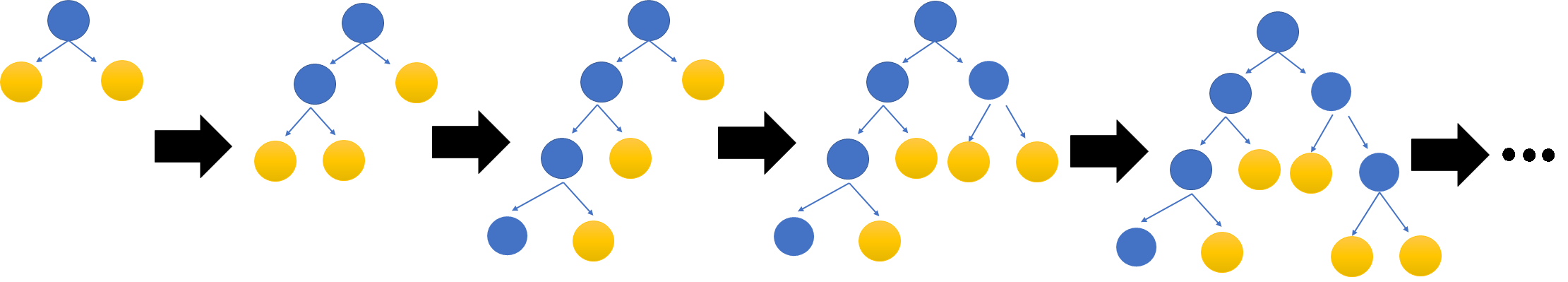}
  \caption{LightGBM uses the leaf-wise tree growth strategy}
  \label{fig:LightGBMGraph}
\end{figure}

The feature space of big data is usually sparse, that is many features are mutually exclusive i.e., they never take nonzero values simultaneously. We can reduce the number of features by combining the mutual exclusive features, and put them into a single feature, called exclusive feature bundle. Therefore, the computational complexity of the histogram-based algorithm can be improved if feature histograms are built from the exclusive feature bundles. This technique is called Exclusive Feature Bundling (EFB) and is employed in LightGBM \cite{36}. Unlike XGBoost, which uses the level-wise tree growth strategy LightGBM uses the leaf-wise tree growth strategy (see Figure \ref{fig:LightGBMGraph}) \cite{36,shi2007best}.

CatBoost is another gradient boosted decision trees (GBDT) that handles categorical features by using one-hot-encoding during training process instead of the preprocessing phase. CatBoost uses the histogram-based algorithm for split searching \cite{37,37-1}. Once again, let $D=\big\{(x_i,y_i) \big | \text{ }x_i \in \mathbb{R}^m, \text{ }y_i \in \mathbb{R}, \text{ } |D|=n\big\}$ be our observed training data set, where $\big\{(x_i,y_i)\big\}_{i=1}^n$ are independent and identically distributed according to some unknown distribution $P$. To train the model, we need to minimize the expected loss,
\begin{equation}\label{eq10}
\text{arg}\min_{F: \mathbb{R}^m \to \mathbb{R}}\mathcal{L}(F)=\text{arg}\min_{F: \mathbb{R}^m \to \mathbb{R}}\mathbb{E}\Bigg(L(y, F(x))\Bigg),
\end{equation} 
where $L$ is a smooth loss function that measures the difference between the prediction $y_i$ and the target $\hat{y}_i$. Like the other two GBDT models, XGBoost and LightGBM, CatBoost builds iteratively a sequence of base predictor, decision trees, to approximate a solution to the optimization problem (\ref{eq10}). Given initial approximation $F^0:\mathbb{R}^m \to \mathbb{R}$, CatBoost constructs iteratively a sequence of approximations $F^k:\mathbb{R}^m \to \mathbb{R}$ for $k=1, 2, \dots$ using the recursive formula,
\begin{equation}
F^k=F^{k-1}+\alpha h^k,
\end{equation}
where $\alpha$ is a step size, and $h^k: \mathbb{R}^m \to \mathbb{R}$ is a binary decision tree that can be computed by 
\begin{equation}
h^k=\text{arg}\min_{h }\mathcal{L}\big(F^{k-1}+ h\big)=\text{arg}\min_{h }\mathbb{E}\Bigg(L\big(y, F^{k-1}(x)+ h(x)\big)\Bigg).
\end{equation}
Note that it is known that the negative gradient is the steepest descent direction (see Line-search methods in \cite{4}), and so $h^k(x)$ can be chosen in a way that it approximates $-g^k(x,y)=-\frac{\partial L(y,s)}{\partial s}$, where $s=F^{k-1}(x)$.

CatBoost uses ordered boosting, a permutation-driven of standard gradient boosting algorithm, in which the model is trained on a subset of data while residuals is calculated on another subset to avoid overfitting. Unlike XGBoost and LightGBM, CatBoost builds balanced trees, which are binary trees in which the height of the left and right subtree of any node differs by not more than one \cite{37,37-1}. 

In the next section, after we prepare the data sets for ML analysis, we train XGBoost, LightGBM and CatBoost on the data sets, and compare their performances. To find the top predictors and find out the relationship between some underlying medical conditions and COVID-19 susceptibility, tree-based feature selection methods are employed for robust (with very good performance) GBDT models.

\section{Methods}\label{section3}
\subsection{Data Resources}
Due to the copious number of topics considered in this paper, four datasets are utilized to produce an accurate, reliable, and comprehensible analysis for the purpose of suggesting plausible conclusions for changes in the COVID-19 case and death rates. 

The first dataset that is used in this paper borrowed from a plethora of databases including the Center of Disease Control (CDC), World Health Organization (WHO), Health Resources and Services Administration (HRSA), Kaiser Family Foundation (KFF), and the Mayo Clinic \cite{21}. The dataset provides a more general outline of the effect of several underlying health conditions, age, gender, and recordings of first and second dose COVID-19 vaccines by each state over the years 2019 through 2021. In terms of the units used for analysis, the regressors used for prediction are percentages and averages of the total population by each state. 

Since the first dataset is objectively small and contained several missing values due to the lack of up-to-date data available on the new COVID-19 variants (Delta and Omicron), this paper also analyzes three larger datasets to produce more reliable and robust results. The three datasets used, AH Provisional COVID-19 Deaths by Educational Attainment, Race, Sex, and Age, AH Monthly Provisional Counts of Deaths by Age Group and HHS region for Select Causes of Death, 2019-2021, and Covid-19 Patient Pre-Condition Dataset, are all publicly available datasets produced by the Center of Disease of Control (CDC) and Kaggle. Note that the second dataset that is analyzed in this paper is constructed from smaller datasets that can be found in \cite{22,23,24,25,26,27,28,29,30}, and the third dataset, Education and Covid, borrowed from \cite{31}.

Unlike other studies investigating COVID-19, this paper seeks to establish a relationship between the high COVID-19 death rate in the Hispanic race and significant underlying health conditions affecting the severity of the symptoms associated with COVID-19. As such, this paper considers a Mexican dataset published by Kaggle outlining patient cause of death over the past couple years (2019–2021) \cite{32}. Therefore, this paper can confidently draw unique and unconventional conclusions about possible reasons for changes in the total case and death count due to COVID-19. Moreover, with multiple datasets, the conclusions made in this paper become more reliable in terms of suggesting possible reasons for changes in the COVID-19 case and death rate. 
For the sake of clarity, below is the list of four datasets, and their details.

\begin{itemize}
\item \textbf{The Education and COVID data:} 
Concerning the CDC dataset for provisional statistics on the selected causes of death by region of the United States, the title published on the Center for Disease Control’s website is stated to be \textit{Ah Provisional Covid-19 Deaths by Educational Attainment, Race, Sex, and Age}. Again, the source for this publicly available dataset was a sub-department within the CDC: the National Center for Health Statistics. This dataset can be found at \textit{ data.cdc.gov/NCHS/AH-Provisional-COVID-19-Deaths-by-Educational-Atta/3ts8-hsrw}\cite{31}. 

Because the data collection process is identical to that of the dataset \textit{Ah Monthly Provisional Counts of Deaths by Age Group and HHS Region for Select Causes of Death, 2019-2021}, the informed consent guidelines should also be considered in determining the credibility of the National Center for Health Statistics’ methodological protocols. On the National Center for Health Statistics’ webpage published by the CDC, the description of what the organization defines to be proper informed consent. In summary, the National Center for Health Statistics states that throughout the informed consent process, survey participants are assured that data collected will be used only for the outlined purposes. Further, the data will not be disclosed or released to others without the consent of the individual. This dataset is in compliance with section 308(d) of the Public Health \textit{Service Act (42 U.S.C. 242m)}, demonstrating its credibility. Additionally, more information about the data collection process and guidelines on informed consent that the National Center for Health Statistics used in this dataset can be found at \textit{ https://www.cdc.gov/nchhstp/programintegration/sc-standards.htm} \cite{31}. 
\item \textbf{The Region-Health-Conditions-COVID data:} 
Concerning the CDC dataset for provisional statistics on the selected causes of death by region of the United States, the title published on the Center for Disease Control’s website is stated to be \textit{Ah Monthly Provisional Counts of Deaths by Age Group and HHS Region for Select Causes of Death, 2019-2021}. While assumed to be the CDC, the source for this publicly available dataset was actually a sub-department within the CDC: the National Center for Health Statistics. This dataset can be found at \textit{data.cdc.gov/NCHS/AH-Monthly-Provisional-Counts-of-Deaths-by-Age-Gro/ezfr-g6hf} \cite{31e}.

In terms of the data collection process used by the National Center for Health Statistics, this organization recorded the stated causes of death by region from weekly death certificates published by the physician who was caring for the specified patient. Once the physician reported the cause of death on the patient’s death certificate, the state in which the patient perished in would send the total number of death certificates in a particular week to the National Center for Health Statistics. Then, the National Center for Health Statistics would produce statistics for cause of death. One particular aspect of the data collection process that makes this methodological approach credible is the strictness in terms of the accuracy and completeness of the death certificate, completeness being, “describing a clear chain of events from the immediate to the underlying cause of death, reporting any other conditions that contributed to death, and providing information that is specific.” Furthermore, to even further demonstrate their reliability, the Natural Center for Health Statistics provides a training module on how to properly determine a cause of death: \textit{https://www.cdc.gov/nchs/nvss/training-and-instructional-materials.htm}. In all, the CDC, while known and accepted as a credible, respected, and professional institution, the guidelines  National Center for Health Statistics use in their data collection process demonstrates the intricacy that this organization values when publishing public data \cite{31e}. 
\item \textbf{The Mexican COVID data:}
With respect to the Mexican dataset this paper considers, the stated title that the Mexican Government has published for the data is \textit{COVID-19 Patient Pre-Condition Dataset}. Additionally, concerning the specific source of the dataset with the Mexican Government that was responsible for compiling this publicly available dataset, the General Directorate of Epidemiology recorded the individual data for each patient considered. While several other studies and papers have referred to or used this dataset in their analyses, one particular study on Kaggle provided the data dictionary and CSV file for public use of the dataset, the website being \textit{www.kaggle.com/tanmoyx/covid19-patient-precondition-dataset} \cite{32}. 

Concerning the ethical and methodological portions of this dataset, the Mexican Government does claim to adhere to the “Open Data” guidelines articulated by the Decree published in the Official Gazette of the Federation on February 20, 2015. This particular decree the department of epistemology is referring to establishes strict criteria for data collection and methodological strategies that can only be used in the “Open Data” policy. For example, especially pivotal to the reliability and ethics of this dataset, informed consent was confirmed to be upheld by the department of epistemology during the collection process. More specifically, the dataset is in accordance with the federal law on protection of personal data held by private parties stating, “Consent: Expression of the will of the data owner by which data processing is enabled.” In all, while considered a foreign dataset, the credibility and ethics are both demonstrated by the precise and strict protocols that the General Directorate of Epidemiology followed during the data collection process for the \textit{COVID-19 Patient Pre-Condition Dataset}. The Law the General Directorate of Epidemiology abided by during the data collection process can be found at \textit{https://www.duanemorris.com/site/static/Mexico-Federal-Protection-Law-Personal-Data.pdf}\cite{32}.
\item \textbf{The COVID-19 data:}  

With regards to the final dataset considered in this paper, the title posted by a sub-department in the Center for Disease Control was stated as a \textit{Comparative Effectiveness of Moderna, Pfizer-Biontech, and Janssen (Johnson \& Johnson) Vaccines in Preventing COVID-19 Hospitalizations among Adults without Immunocompromising Conditions - United States, March–August 2021}. While considered a study published in the Morbidity and Mortality Weekly Report (MMWR), this paper utilizes the dataset on vaccine efficacy that this study considered as well. Furthermore, this paper also includes other relevant confounding variables to analyze the efficacy of the COVID-19 booster, the Pfizer, Johnson \& Johnson, and Moderna vaccines such as median age, gender, and several underlying health conditions \cite{31ee}. 

	Concerning the methodology used to compile this dataset, the MMWR early release only summarized and analyzed this dataset. On the other hand, the CDC recorded the total number of vaccinations by type released by each state in the specified months March through August of 2021. As such, the CDC used a variety of new IT systems that support vaccine logistics and administration, some of these new systems being the Vaccine Tracking System, Immunization Information Systems, Vaccine.gov, and the Immunization (IZ) Gateway. Concerning the data collection systems used, the CDC uses three new systems for data reporting on COVID-19 vaccination by state: COVID-19 Data Clearinghouse, IZ Data Lake, and Privacy-preserving record linkage (PPRL). With regards to informed consent being administered and upheld, the CDC utilize the same protocols when formulating the datasets \textit{Ah Monthly Provisional Counts of Deaths by Age Group and HHS Region for Select Causes of Death, 2019-2021 and Ah Provisional Covid-19 Deaths by Educational Attainment, Race, Sex, and Age} \cite{31ee}.

\end{itemize}
\subsection{Data Preprocessing and Python Codes Resources}
\subsubsection{The Education and COVID data preprocessing and Python codes}
The dataset contains 72 rows and 8 columns with no missing value. But we add a new column namely $\text{CTDPercentage}=\frac{\text{COVID-19 Deaths}}{\text{Total Deaths}} \times 100\%$ (Figure \ref{fig:3} displays columns and the first five rows). Python codes corresponding to the analysis of this dataset is available in Section 1 (\textit{Analyzing the Education and COVID data}) of the paper Github repository \cite{33}.  
\begin{figure}[H]
  \includegraphics[width=.9\linewidth]{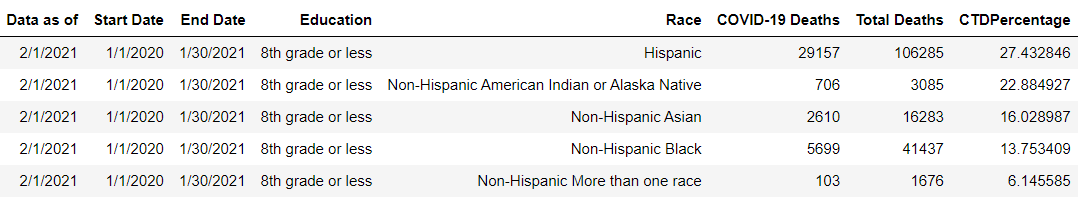}
  \caption{The columns and the first five rows of the third dataset}
  \label{fig:3}
\end{figure}

\subsubsection{The Region-Health-Conditions-COVID data preprocessing and Python codes}
\begin{figure}[H]
  \includegraphics[width=.7\linewidth]{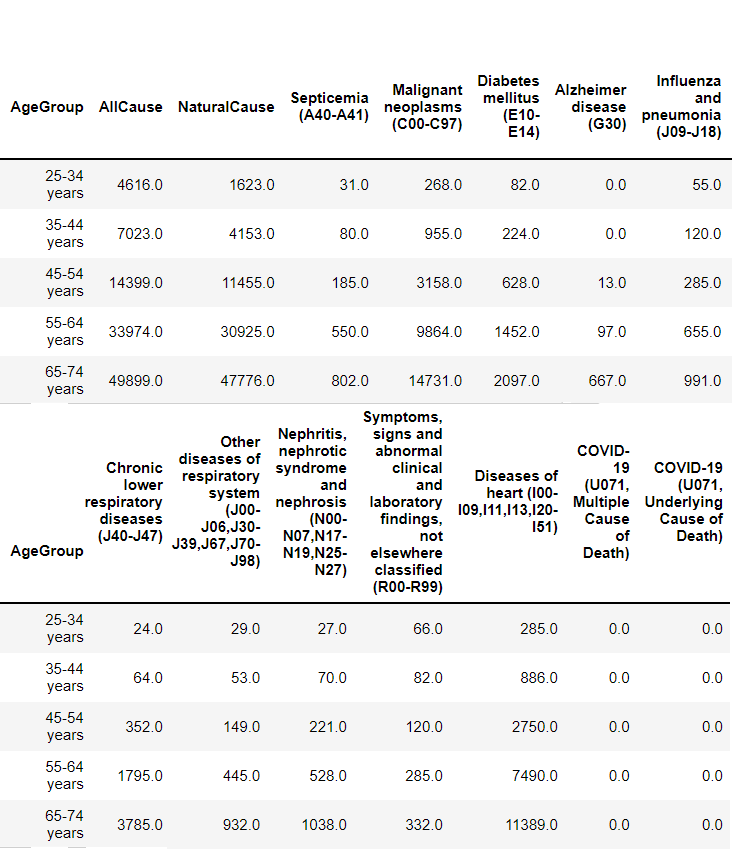}
  \caption{The columns and the first five rows of the second dataset}
  \label{fig:2}
\end{figure}

The dataset contains 3410 rows and 36 columns with 53240 missing values. However, the original dataset contains several unnecessary columns for our purpose. Thus, we remove all unnecessary columns, and all rows with missing values, and as the result, we end up with a clean tabular dataset containing 1252 rows and 15 columns (Figure \ref{fig:2} displays columns and the first five rows).  But, this paper uses the following variables as main variables for analysis: AllCause (the death rate from all causes of death for a population in each time period), AllNatural (internal factors-like a medical condition or a disease- as opposed to external factors, like trauma from an accident), Septicemia, diseases of the heart, and diabetes mellitus. These factors are considered as primary effects for predicting the total COVID-19 death count because of their preconceived connection to COVID-19 as well as the extremity of their symptoms. Python codes corresponding to the analysis of this dataset is available in Section 3 (\textit{Analyzing the Region-Health-Conditions-COVID data}) of the paper Github repository \cite{33}.  

\subsubsection{The Mexican COVID data preprocessing and Python codes}
The dataset is already encoded to numbers. There exists a Catalog attached to the dataset for decoding the categories under each column. We consider “cov-res”, which stands for COVID-Result, as our target variable. In the original dataset, this variable contains 3 categories, 1: Positive SARS-CoV-2, 2: Non-positive SARS-CoV-2 and 3: Pending result. But, since the category 3, Pending result, cannot help us train machine learning models, we remove all rows with label 3, Pending result, for target variable. As the result, we end up with a clean tabular dataset containing 499692 rows and 23 columns. Python codes corresponding to the analysis of this dataset is available in Section 4 (\textit{Analyzing the Mexican covid Data}) of the paper Github repository \cite{33}.  
\begin{figure}[H]
  \includegraphics[width=.9\linewidth]{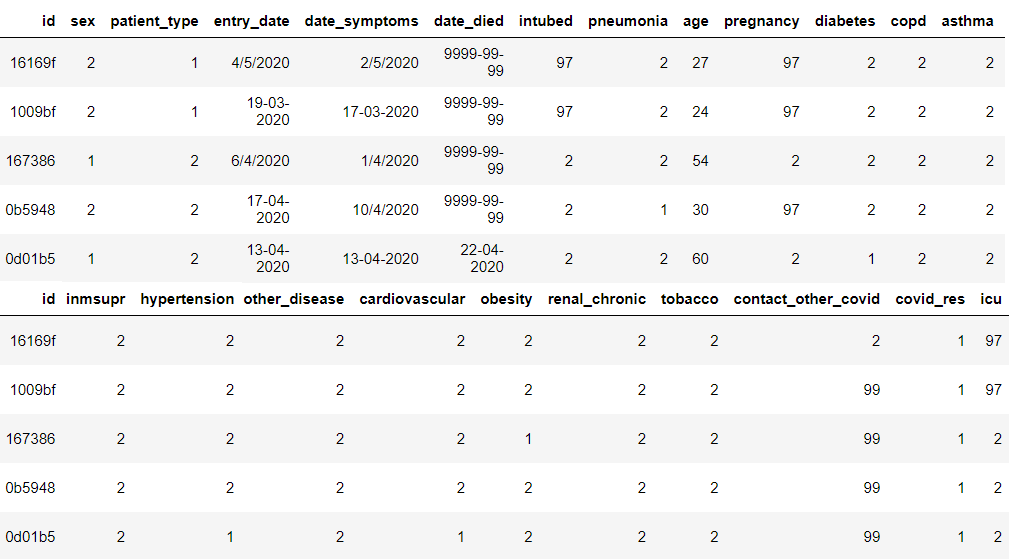}
  \caption{The columns and the first five rows of the fourth dataset}
  \label{fig:3}
\end{figure}

\subsubsection{The COVID-19 data preprocessing and Python codes}
\begin{figure}[H]
  \includegraphics[width=.8\linewidth]{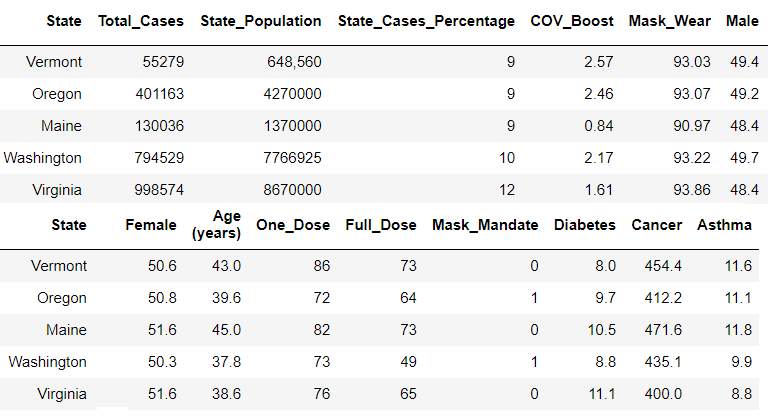}
  \caption{The columns and the first five rows of the first dataset}
  \label{fig:1}
\end{figure}

The dataset contains 24 rows and 13 columns, but since we need to scale the data, we add two columns: State-Population and $\text{State-Cases-Percentage}=\frac{\text{Total-Cases}}{\text{State-Population}} \times 100\%$ (Figure \ref{fig:1} displays columns and the first five rows). As such, the variables this paper will be investigating to address the initial hypothesis include COV-Boost (the percentage of the population who have received the COVID-19 Booster vaccine in the past two weeks), One-Dose (the percentage of the population in each state who has received one dose of either Pfizer or Moderna vaccines), and Full-Dose (the percentage of the population in each state who has received both doses of either Pfizer or Moderna vaccines). Additional variables that will be considered for the purpose of providing a comprehensive review of the effect of the COVID-19 boosters and vaccines on the total case count in the USA include the percentage of the population who wear a mask in the last month, whether a state has a mask mandate or not, gender, age, and some underlying health conditions that are particularly important to consider when discussing the COVID-19 symptoms. Python codes corresponding to the analysis of this dataset is available in Section 2 (\textit{Analyzing the COVID-19 dataset}) of the paper Github repository \cite{33}.  
\subsection{Computational Process}
In this section, we discuss machine learning methods and statistical tests that are utilized in this paper separately for each dataset by following the order of Python codes in the repository \cite{33}.  
\subsubsection{Analyzing the Education and COVID data}
\begin{figure}[H]
  \includegraphics[width=.8\linewidth]{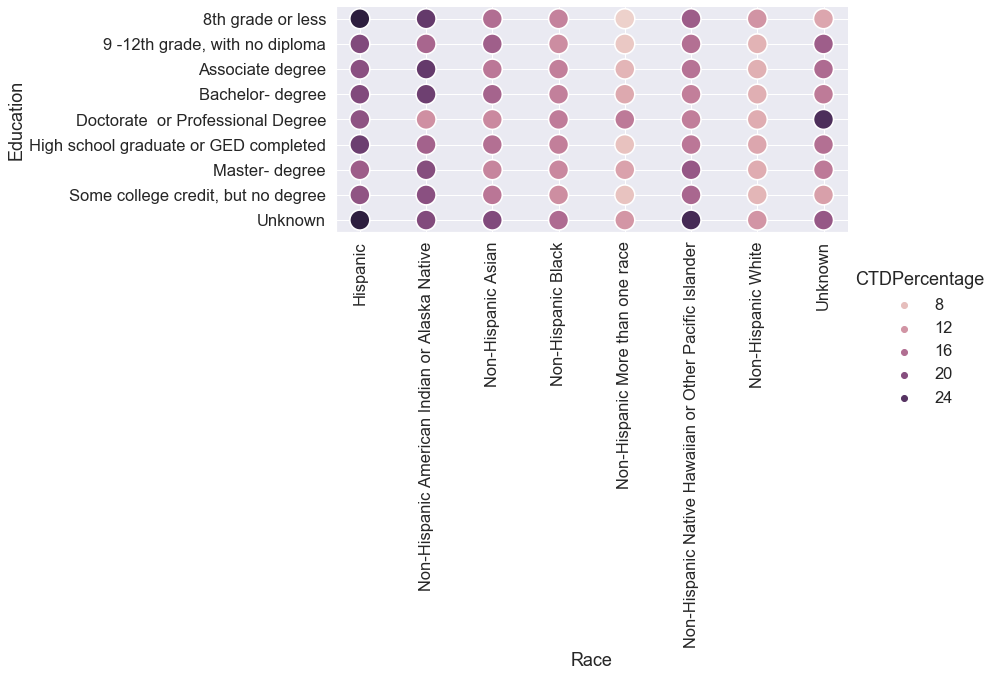}
  \caption{The relationship among COVID-related total deaths percentage, education, and race}
  \label{fig:5}
\end{figure}

The first dataset this paper utilizes to address the questions mentioned in the introduction concerns the relationship between educational attainment, race, and the COVID-19 case rate. In this dataset provided by the CDC, educational attainment ranges from 8th grade or less to general graduate degrees while race considers Hispanics in relation to other subgroups of the non-white, U.S. population. To begin the analysis, this paper provides an overview of the interaction between race and education as a predictor for COVID-19 Total Death Percentage (CTDPercentage) (see Figure \ref{fig:5}). 

\begin{figure}[H]
  \includegraphics[width=\linewidth]{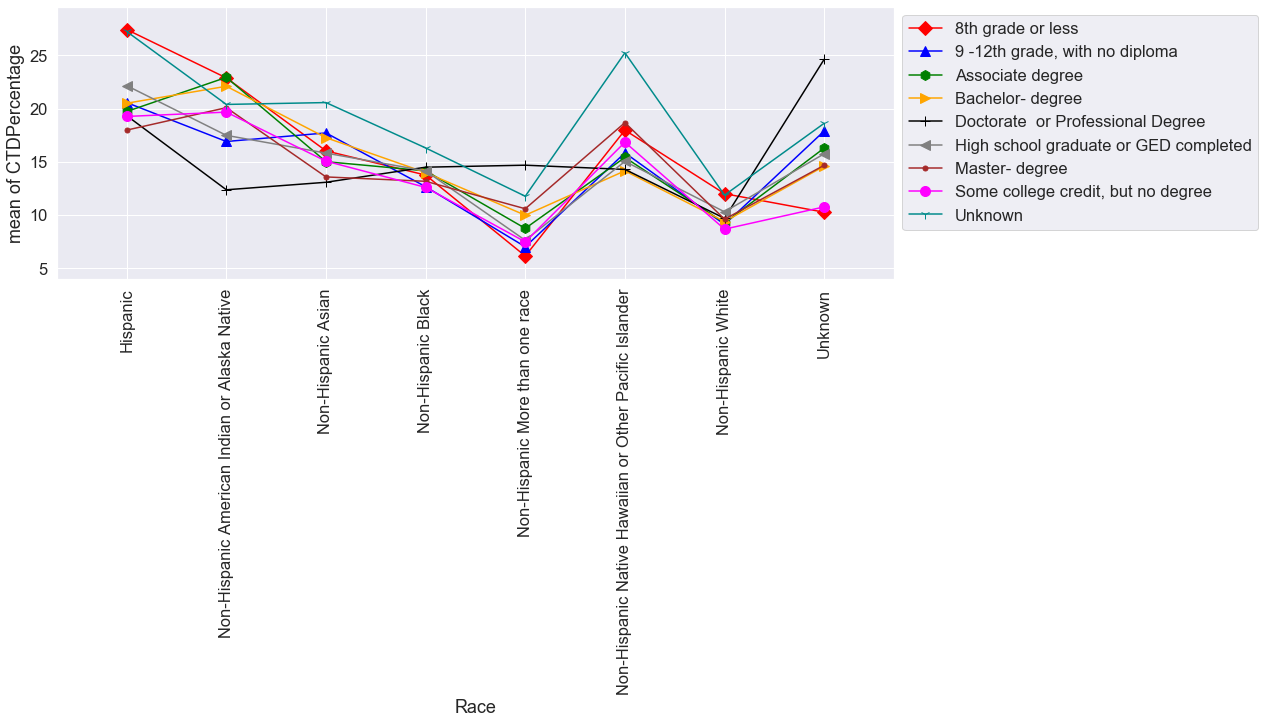}
  \caption{The relationship among COVID-related total deaths percentage, education, and race }
  \label{fig:6}
\end{figure}

\begin{figure}[H]
  \includegraphics[width=\linewidth]{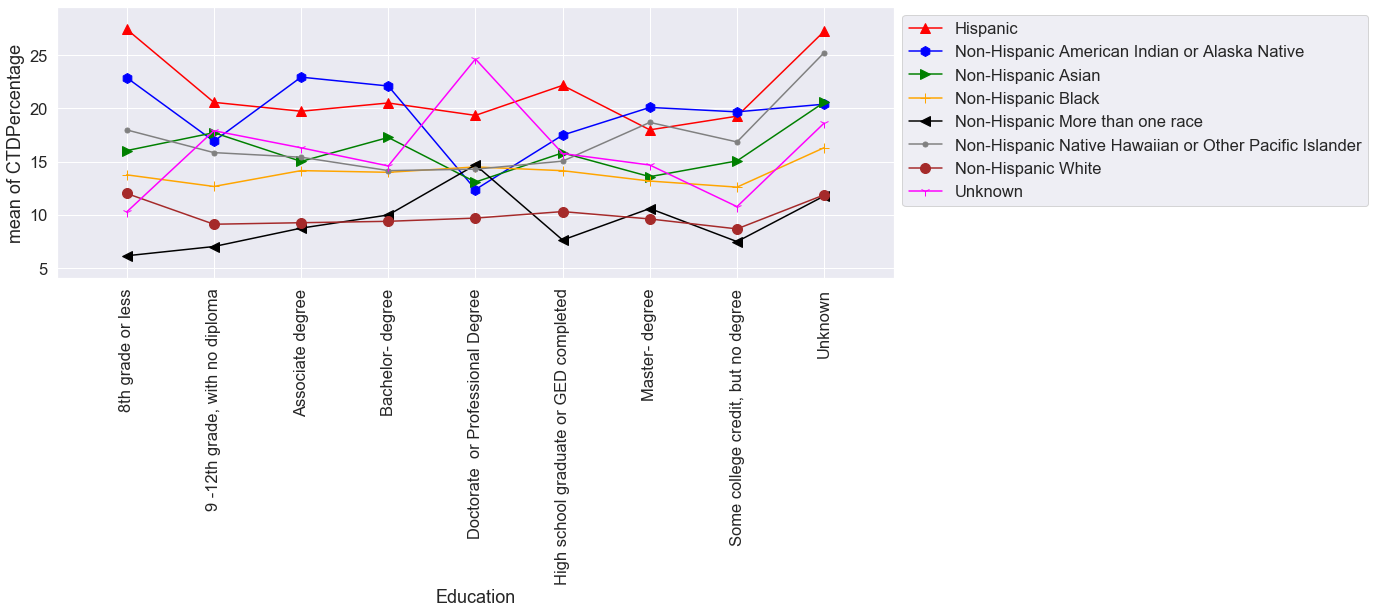}
  \caption{The relationship among COVID-related total deaths percentage, education, and race }
  \label{fig:7}
\end{figure}

In general, in relation to the other ethnic subgroups considered (such as Non-Hispanic Asian or Non-Hispanic Black), the Hispanic subgroup tends to have the highest COVID-19 total death percentage. More specifically, Hispanics with an educational attainment of “unknown” or “8th grade or less” tend to have a CTDPercentage of approximately 24\% or more. Additionally, Hispanics, even with a “Master degree” or a “Bachelor degree” still tend to have a higher CTDPercentage in relation to other non-Hispanic subgroups considered in the dataset. 
 We use interaction plots to show how the COVID-related total deaths percentage (CTDPercentage) depends on Education and Race differences (see Figures \ref{fig:6} and \ref{fig:7}).

\begin{figure}[H]
  \includegraphics[width=.6\linewidth]{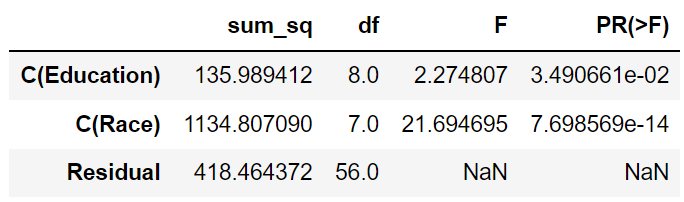}
  \caption{ANOVA is constructed under the assumption that there is no significant interaction between Education and Race}
  \label{fig:8}
\end{figure}

\begin{figure}[H]
    \centering
    \subfloat[Box Plot of COVID Total Death Percentage by Race]{{\includegraphics[width=7cm]{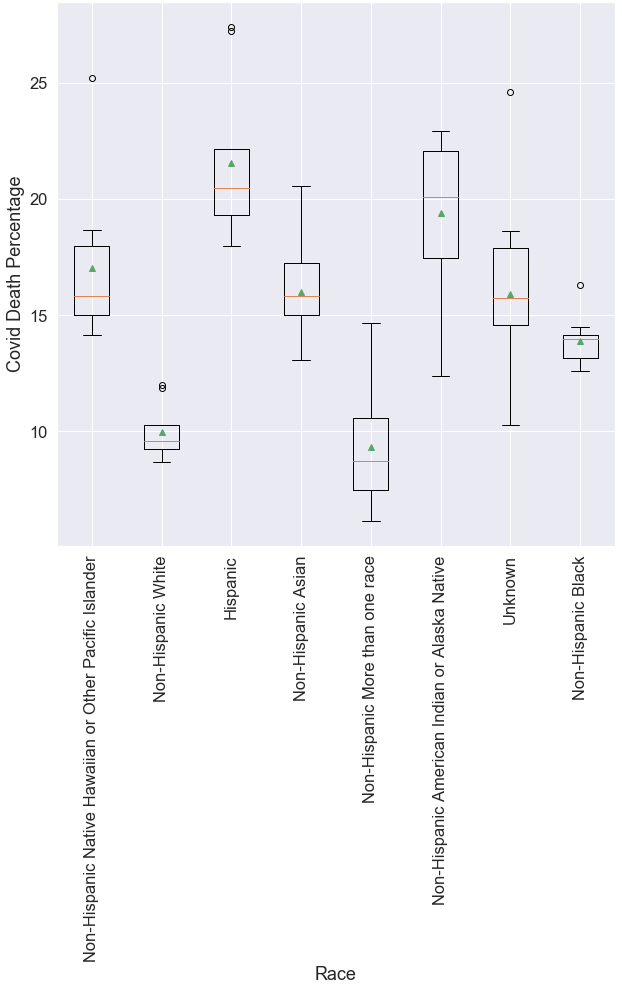} }}
    \qquad
    \subfloat[Box Plot of COVID Total Death Percentage by Education]{{\includegraphics[width=7cm]{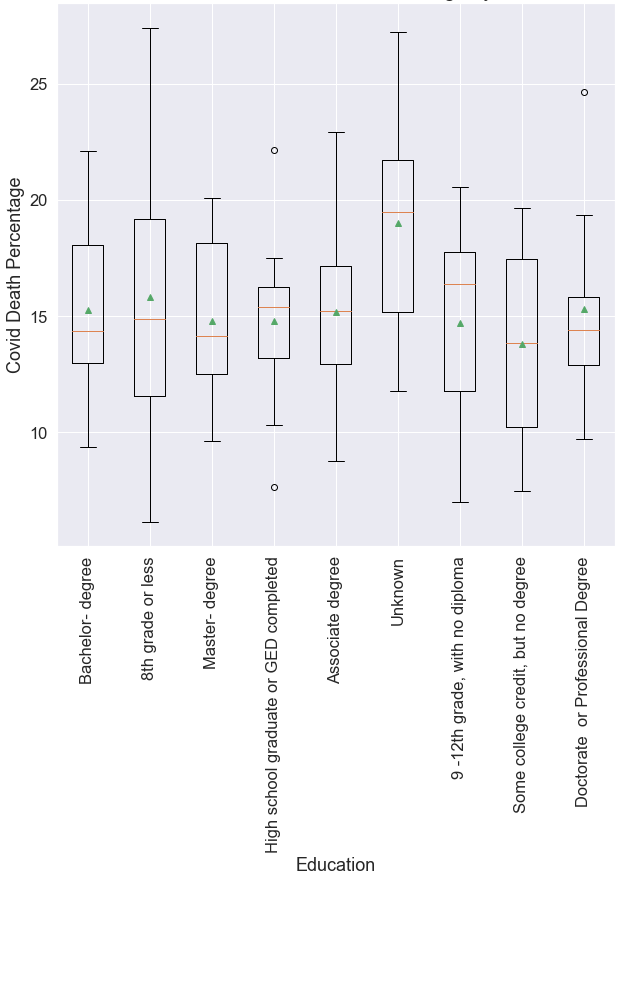} }}
    \caption{Box Plots}
    \label{fig:9}
\end{figure}

To further demonstrate there is a strong relationship between the interaction of educational attainment and the Hispanic race with the COVID-19 total death percentage, this paper considers an Analysis Of Variance (ANOVA) and box plot relating education and race with CTDPercentage. In short, we examine the data to find out whether there are any significant differences between the means of the values of the CTDPercentage for each categorical value of Education and Race. This is something that we can also visualize using a box-plot as well (see Figures \ref{fig:8} and \ref{fig:9}). Note that we assume the interaction between Education and Race is non-significant when the ANOVA is constructed. Moreover, since the predictors, Education and Race, do not have effect on the outcome variable, CTDPercentage, the null hypothesis is evaluated with regard to Pr$>$F, which is p-value associated with the F statistic. 

By Figures \ref{fig:6} and \ref{fig:7}, it is evident that the Hispanic race tends to have the highest COVID-19 total death percentage in relation to other ethnic groups. For example, Hispanics with an educational attainment of 8th grade or less will tend to have a COVID-19 total death percentage between 25\% and 30\%. On the other hand, non-Hispanic blacks with an educational attainment of 8th grade or less will have a COVID-19 total death percentage of approximately 15\%. Therefore, it is evident that there is a relationship between race and total deaths due to COVID-19 apart from educational attainment as a result from analyzing the ANOVA. In general, the box plots (Figure \ref{fig:9}) show that the Hispanic race has a greater median than all other ethnic groups considered in this dataset, approximately a 21\% COVID-19 total death rate. Furthermore, the Hispanic ethic group has the highest first (25\%) and third (75\%) quartiles, demonstrating that many of the observations representing the Hispanic subgroup have higher COVID-19 total death percentages than all other ethnic groups considered. Evident from the results produced by Figures \ref{fig:6}--\ref{fig:9}, this paper subsequently considers two additional USA datasets and one additional Mexican dataset to propose possible reasons for this relationship between the Hispanic race and the COVID-19 total death percentage. 

\subsubsection{Analyzing the Region-Health-Conditions-COVID data}
To better understand the possible reasons for changes in the COVID-19 death count, this paper considers the relationship between underlying health conditions and illnesses and COVID-19 case and death rates.

\begin{figure}[H]
  \includegraphics[width=.9\linewidth]{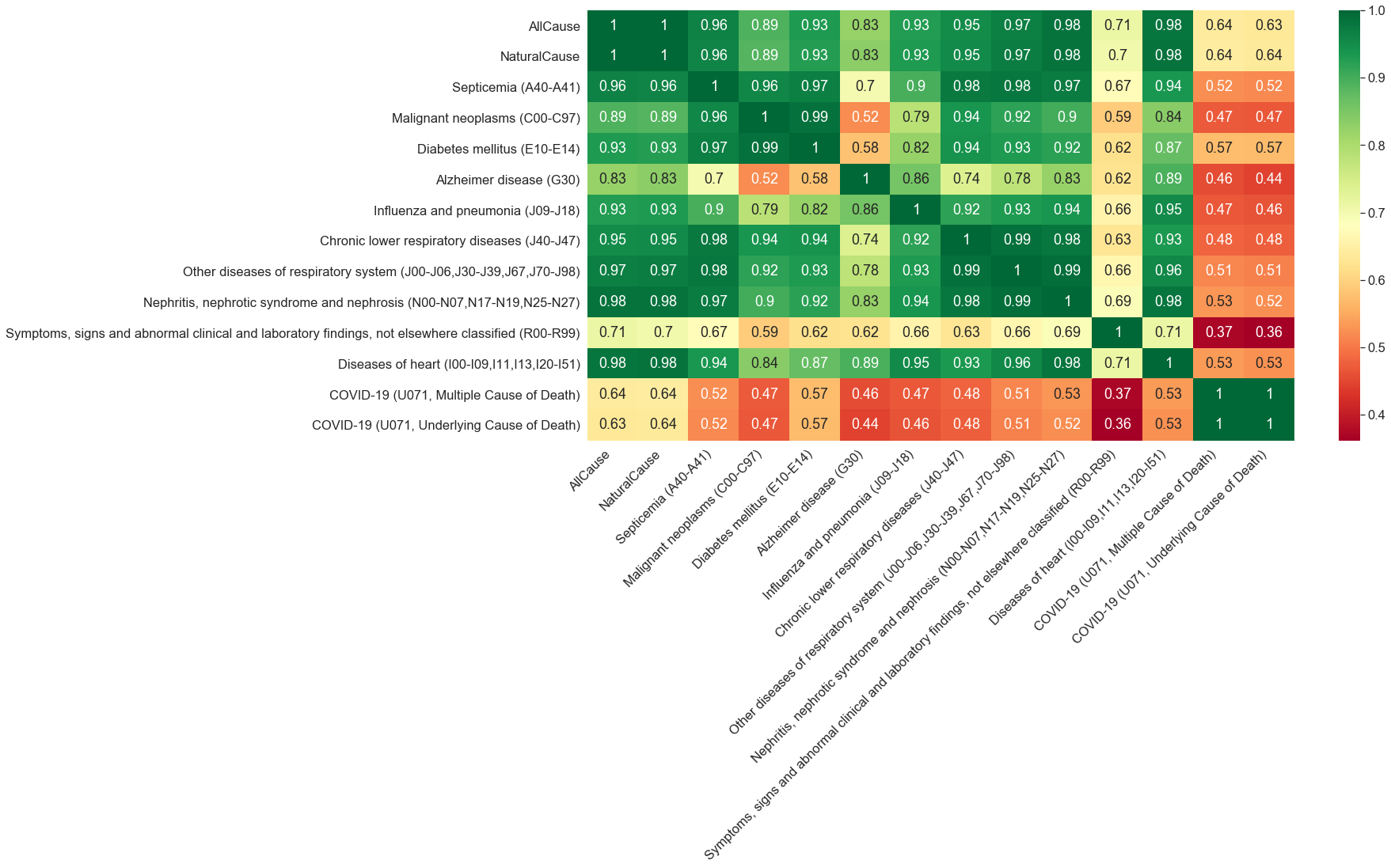}
  \caption{The heat map matrix depicting the correlation and R-squared values for some important variable}
  \label{fig:11}
\end{figure}

\begin{figure}[H]
  \includegraphics[width=.8\linewidth]{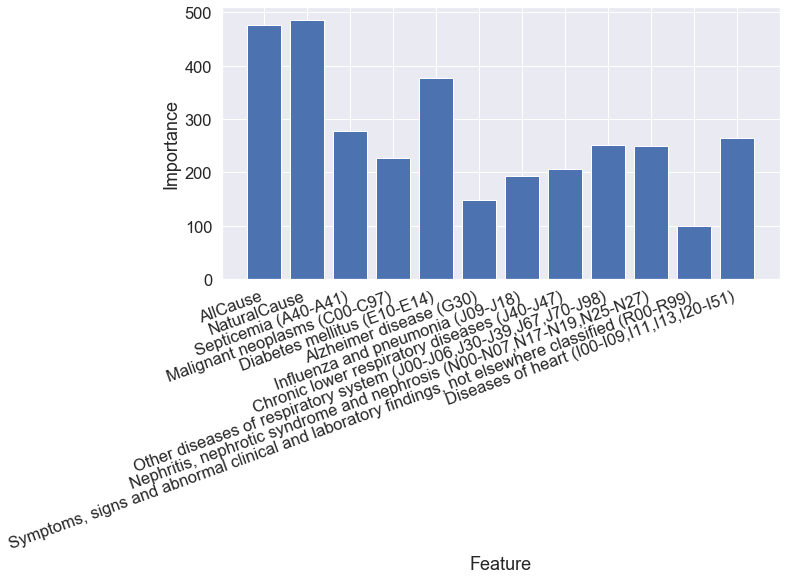}
  \caption{Feature importance scores obtained by regression model}
  \label{fig:12}
\end{figure}

In the dataset Region-Health-Conditions-COVID, several underlying health conditions are considered in relation to region in the United States. To begin the analysis, this paper first provides a correlation, \textit{heat map} matrix depicting the correlation and R-squared values for some important variable included in the dataset (see Figure \ref{fig:11}). Evident from this graphic (Figure \ref{fig:11}), several underlying health conditions are positively correlated with the “multiple” and “single” COVID-19 cause of deaths. More specifically, all-cause mortality, natural cause mortality, and Diabetes Mellitus (E10–E14) have the highest, positive correlation coefficients in relation to all other illnesses and underlying health conditions considered in this dataset. Furthermore, these high, positive correlation values indicate that as all cause, natural cause, or diabetes mellitus cause of deaths increase, then COVID-19 “multiple” and “single” cause of deaths also tend to increase. 
 To further convey the profoundness of these strong, positive relationships between these selected underlying health conditions and COVID-19 cause of death, we split the dataset into two subsets: training set containing 838 datapoints (75\%) and test set containing 414 datapoints (25\%), and train a regression model with \textit{COVID-19 (U071, Underlying Cause of Death)} as the target variable. Figure \ref{fig:12} displays a bar graph measuring feature importance scores of the regression model (the magnitude of “importance” of each underlying health condition or illness when predicting COVID-19 “single” and “multiple” cause of deaths). Evident by the results of the bar graph, all-cause mortality, natural cause mortality, and diabetes mellitus have the largest magnitudes of importance in relation to other underlying health conditions considered. Therefore, it is plausible to conclude that there is a strong relationship between underlying health conditions and COVID-19 deaths, specifically a strong, positive relationship evident from the universally positive correlation coefficients seen in the heat map in Figure \ref{fig:11}.
 \subsubsection{Analyzing the Mexican covid Data}
\begin{figure}[H]
\centering
\begin{subfigure}{.5\textwidth}
  \centering
    \caption{Bar graph grouped: Diabetes versus Covid result}
  \includegraphics[width=.82\linewidth]{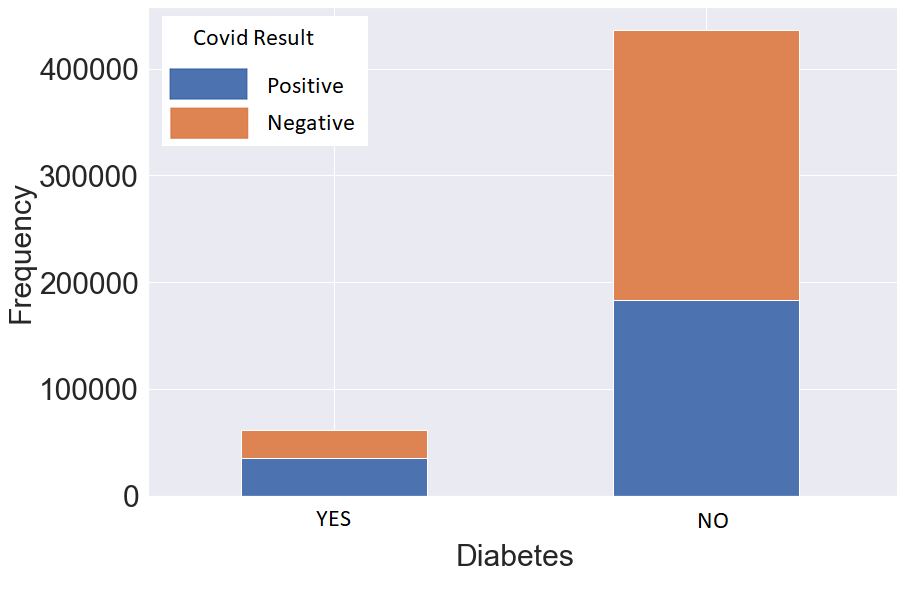}

  \label{fig:sub11}
\end{subfigure}%
\begin{subfigure}{.5\textwidth}
  \centering
    \caption{Bar graph grouped: Asthma versus Covid result}
  \includegraphics[width=.8\linewidth]{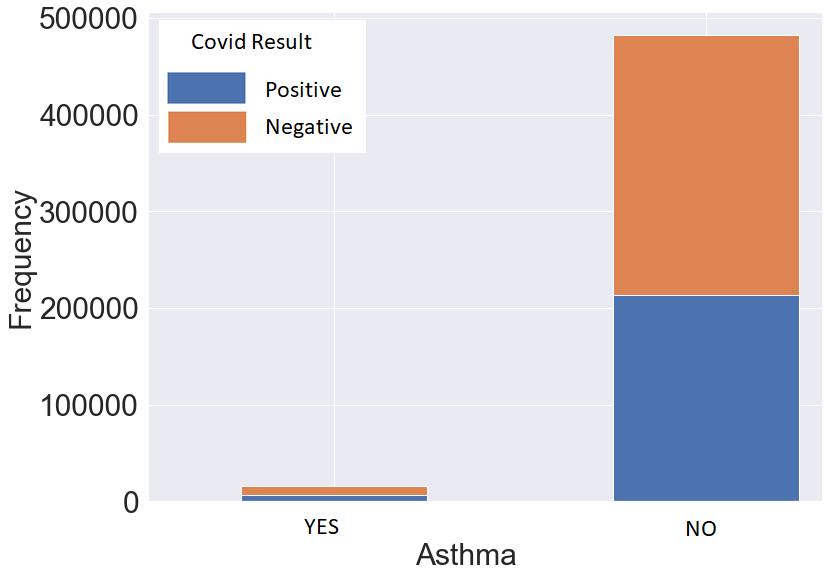}

  \label{fig:sub12}
\end{subfigure}
\caption{Bar graph grouped for the variables that reject the null hypothesis of the Chi-Squared test with significance level $\alpha=0.05$}
\label{fig:test11121}
\end{figure}

\begin{figure}[H]
\centering
\begin{subfigure}{.5\textwidth}
  \centering
    \caption{Bar graph grouped: Cardiovascular versus Covid result}
  \includegraphics[width=.82\linewidth]{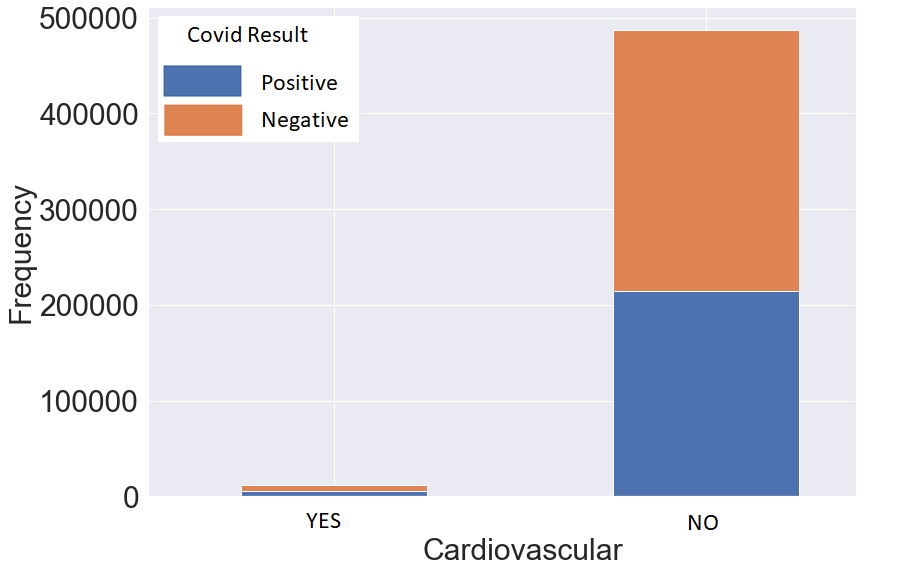}

  \label{fig:sub21}
\end{subfigure}%
\begin{subfigure}{.5\textwidth}
  \centering
    \caption{Bar graph grouped: Hypertension versus Covid result}
  \includegraphics[width=.8\linewidth]{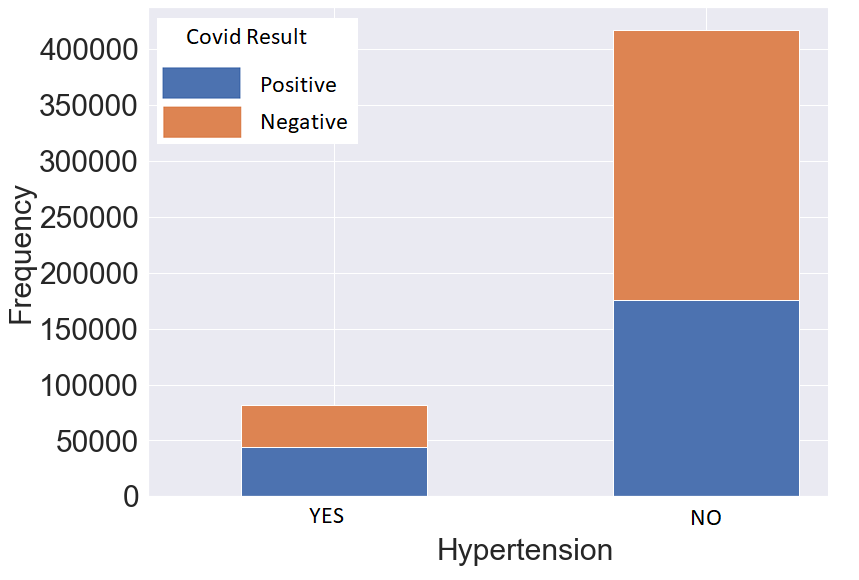}

  \label{fig:sub22}
\end{subfigure}
\caption{Bar graph grouped for the variables that reject the null hypothesis of the Chi-Squared test with significance level $\alpha=0.05$}
\label{fig:test11122}
\end{figure}

\begin{figure}[H]
\centering
\begin{subfigure}{.5\textwidth}
  \centering
    \caption{Bar graph grouped: Renal Chronic versus Covid result}
  \includegraphics[width=.82\linewidth]{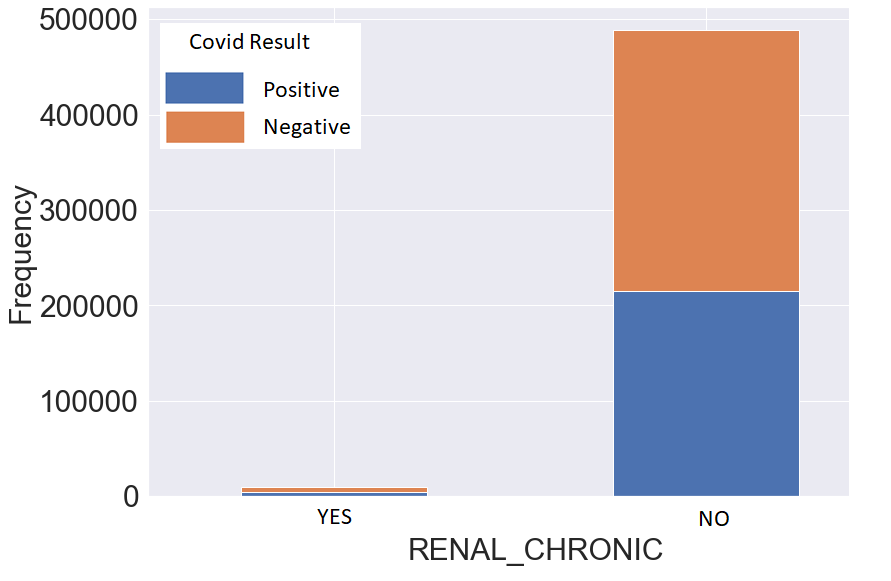}

  \label{fig:sub31}
\end{subfigure}%
\begin{subfigure}{.5\textwidth}
  \centering
    \caption{Bar graph grouped: Tobacco versus Covid result}
  \includegraphics[width=.8\linewidth]{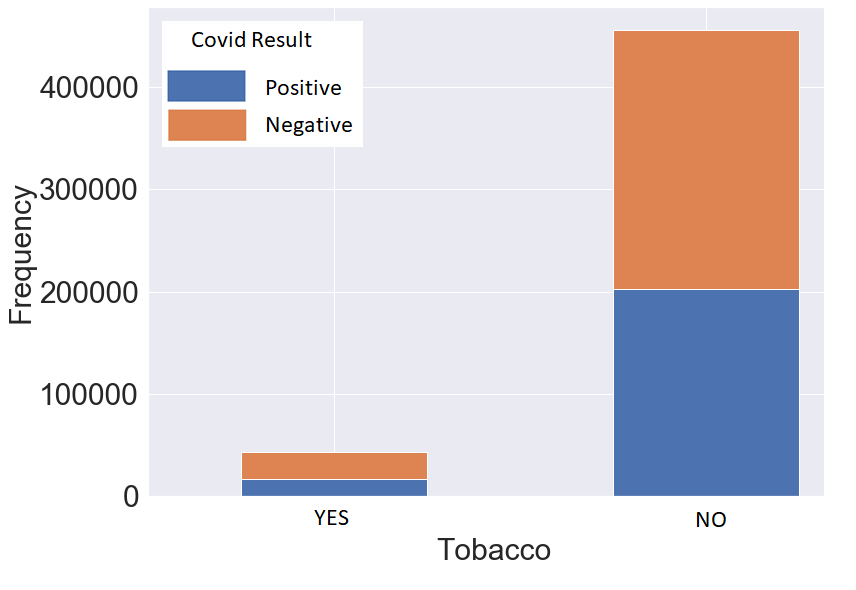}

  \label{fig:sub32}
\end{subfigure}
\caption{Bar graph grouped for the variables that reject the null hypothesis of the Chi-Squared test with significance level $\alpha=0.05$}
\label{fig:test11123}
\end{figure}

 Because of the conclusions made about the relationship between race and COVID-19 deaths and the relationship between underlying health conditions and COVID-19 deaths, this paper also includes analysis from a Mexican patient cause of death dataset published by Kaggle in 2021. Like the previous dataset considered, this dataset provides a complete recording of diagnoses for cause of deaths for several patients admitted to hospitals in Mexico. Since the target variable, Covid-result (``covid-res''), and the majority of input variables are categorical (nominal or ordinal), we first apply Chi-squared tests under the following  hypotheses:
  \[\begin{cases}
                    H_0: \text{Covid-result (``covid-res'') is independent of variable $i$}, \\
                     H_1: \text{Covid-result (``covid-res'') is not independent of variable $i$},
\end{cases} \]
where variable $i$ is one of the input variables.  By a significance level of $\alpha=0.05$,  it turns out that the null hypothesis $H_0$ is rejected with a p-value less than $0.001$ for diabetes, asthma, cardiovascular, hypertension, renal-chronic and tobacco (see Figures \ref{fig:test11121}--\ref{fig:test11123}). So, there also appears to be a strong relationship between COVID-19 cause of deaths and underlying health conditions. As such, to hone in on this reoccurring relationship, we utilize XGBoost \cite{34,35}, LightGBM \cite{36}, and CatBoost \cite{37} to find the most related factors (features) to the Covid result. 

Gradient boosting is a machine learning technique that converts weak learners into strong learners. XGBoost \cite{34,35}, LightGBM \cite{36}, and CatBoost \cite{37} are decision-tree based ensemble supervised learning algorithms that follow the principle of gradient boosting. Using One-Hot-Encode, CatBoost is specifically designed for categorical data, and usually has a better performance on it in comparison with XGBoost and LightGBM. All LightGBM, XGBoost, and CatBoost have a list of tunable hyperparameters that affect learning and eventual performance, but it is often not clear which will perform best until testing them all. These three boosting algorithms are compared with each other in \cite{38}. 

\begin{figure}[H]
\centering
\begin{subfigure}{.33\textwidth}
  \centering
    \caption{CatBoost with 10 trees}
  \includegraphics[width=\linewidth]{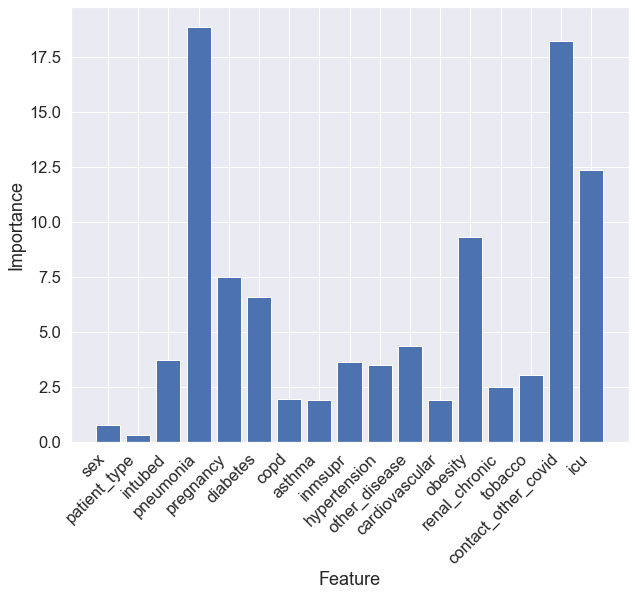}
  \label{fig:cat10}
\end{subfigure}%
\begin{subfigure}{.33\textwidth}
  \centering
    \caption{CatBoost with 100 trees}
  \includegraphics[width=.98\linewidth]{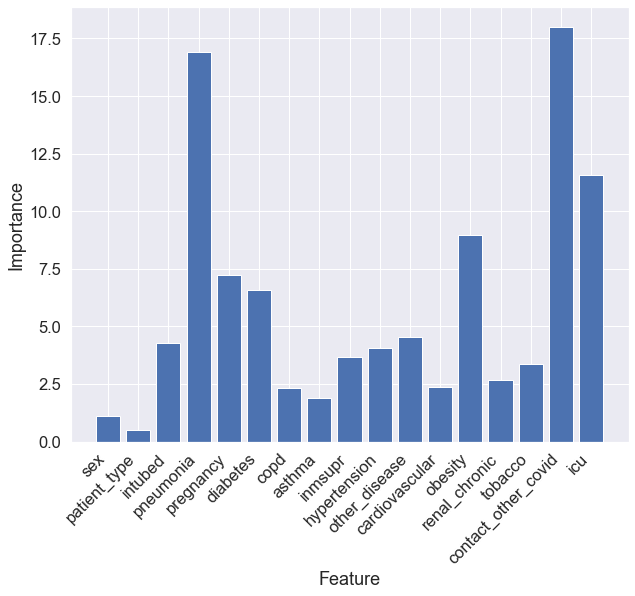}
  \label{fig:cat100}
\end{subfigure}
\begin{subfigure}{.33\textwidth}
  \centering
    \caption{CatBoost with 1000 trees}
  \includegraphics[width=.98\linewidth]{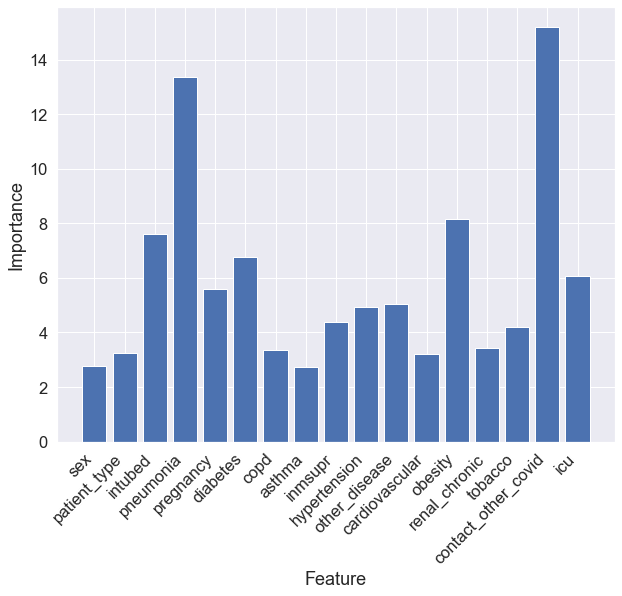}
  \label{fig:cat1000}
\end{subfigure}
\caption{Feature importance scores for CatBoost with 10, 100, and 1000 trees}
\label{fig:catwith}
\end{figure}

\begin{figure}[H]
\centering
\begin{subfigure}{.33\textwidth}
  \centering
    \caption{CatBoost with 10 trees}
  \includegraphics[width=\linewidth]{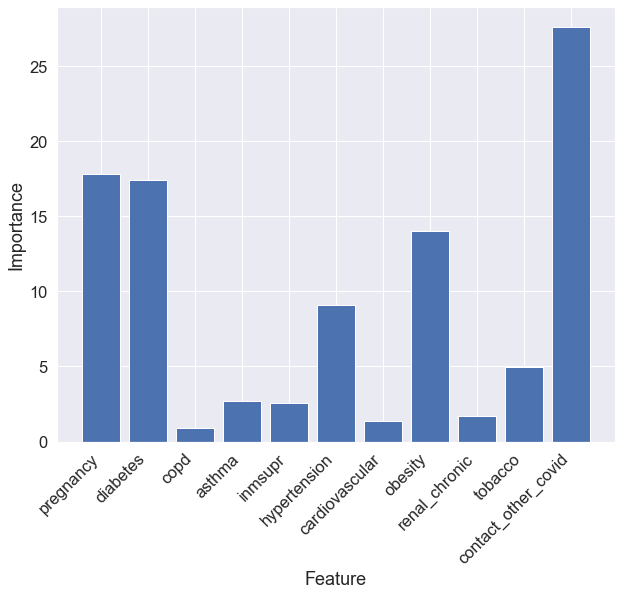}
  \label{fig:cat10w}
\end{subfigure}%
\begin{subfigure}{.33\textwidth}
  \centering
    \caption{CatBoost with 100 trees}
  \includegraphics[width=.98\linewidth]{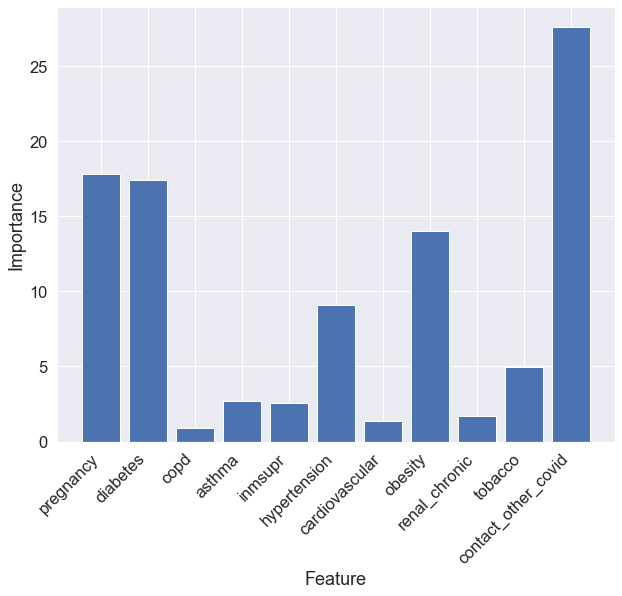}
  \label{fig:cat100w}
\end{subfigure}
\begin{subfigure}{.33\textwidth}
  \centering
    \caption{CatBoost with 1000 trees}
  \includegraphics[width=.98\linewidth]{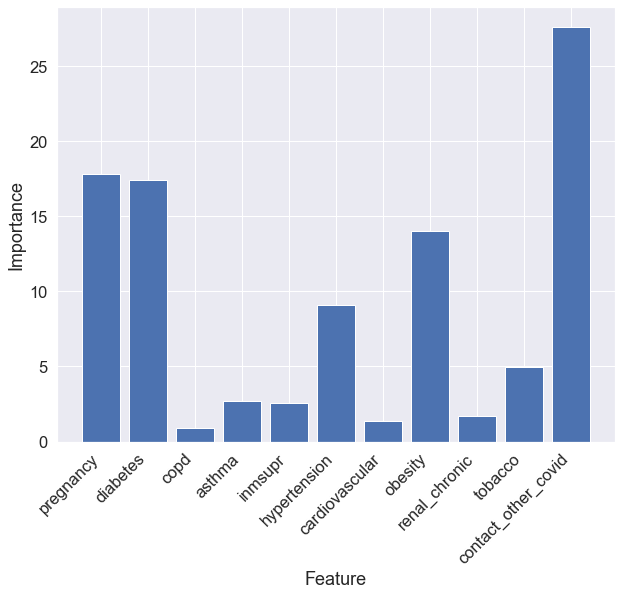}
  \label{fig:cat1000w}
\end{subfigure}
\caption{Feature importance scores for CatBoost with 10, 100, and 1000 trees on modified dataset}
\label{fig:catwithout}
\end{figure}

\begin{figure}[H]
  \includegraphics[width=.5\linewidth]{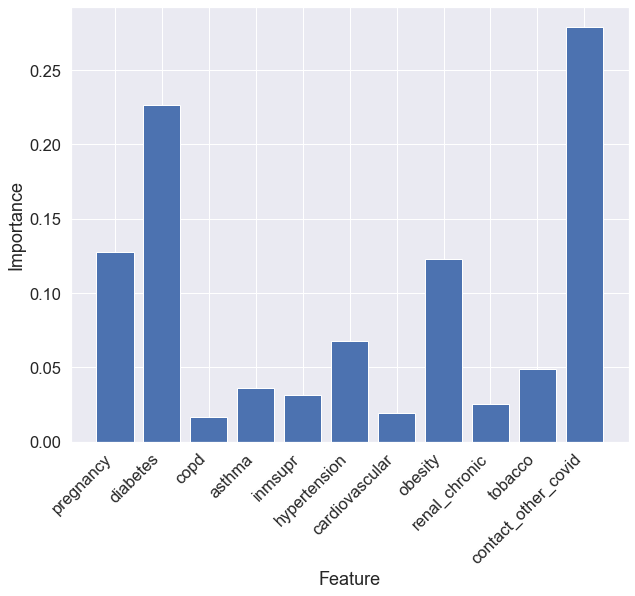}
  \caption{The obtained feature importance scores by XGBoost}
  \label{fig:XGBoost}
\end{figure}

\begin{figure}[H]
  \includegraphics[width=.5\linewidth]{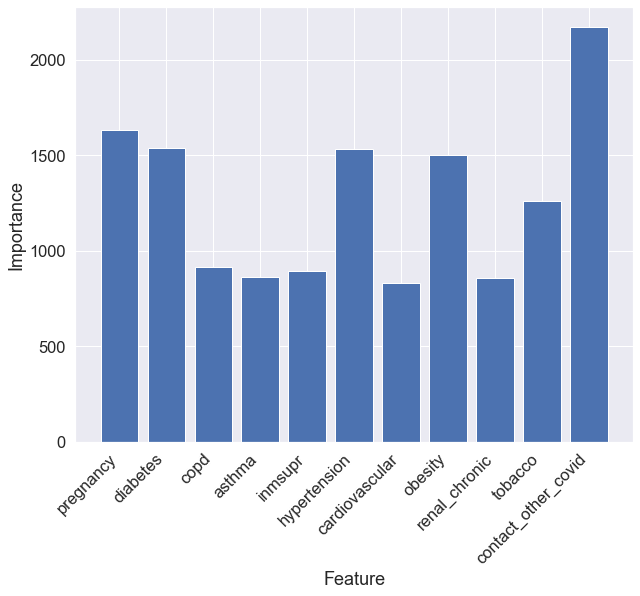}
  \caption{The obtained feature importance scores by LightGBM}
  \label{fig:LightGBM}
\end{figure}

We first train CatBoost algorithms with 10, 100 and 1000 trees using input variables \textit{sex, patient-type, intubed, pneumonia, pregnancy, diabetes, copd, asthma, inmsupr, hypertension, other-disease, cardiovascular, obesity, renal-chronic, tobacco, contact-other-covid} and \textit{icu} and the target variable \textit{cov-res} (Covid result). Figure \ref{fig:catwith} displays feature importance scores of CatBoost models with 10, 100 and 1000 trees. 
Figure \ref{fig:catwith} indicates that some variables such as \textit{intubed, icu} and \textit{ pneumonia} have a positive strong correlation with the target variable \textit{cov-res}. Since this relationship is obvious and does not help us have a better understanding about other features, we remove these features and retrain CatBoost using variables associated with underlying medical conditions as inputs. Figure \ref{fig:catwithout} displays feature importance scores of CatBoost models with 10, 100 and 1000 trees on the modified dataset. Figure \ref{fig:catwithout} indicates that CatBoost feature importance scores are almost identical regardless of the number of trees. 
To explore the correlation between the Covid result and underlying medical conditions further, we train XGBoost on the modified dataset containing only input variables associated with the underlying medical conditions. Figure \ref{fig:XGBoost} displays the obtained feature importance scores.  
Figure \ref{fig:LightGBM} also displays the feature importance scores obtained by a LightGBM with 2000 trees (maximum depth equals 10).
 
In the results obtained by XGBoost, LightGBM and CatBoost, Diabetes appears to have strong importance when predicting the cause of death due to COVID-19, implying that diabetes could make the symptoms associated with COVID-19 more severe and fatal. Additionally, Mexican patients with conditions such as hypertension, obesity, and even tobacco overuse issues, also appear to be the most important features for predicting COVID-19 cause of death. As a result, of the high importance of these features, it can be concluded that although there tends to be a strong relationship between race and the COVID-19 total death percentage, the effect of underlying health conditions and illnesses appears to be strongly influencing and contributing to this relationship rather than educational attainment alone. Therefore, it is possible that the Hispanic subgroup considered in the first dataset could contain higher than normal concentrations of underlying health conditions, ultimately explaining the high COVID-19 Total Death Percentages for this subgroup.
\subsubsection{Analyzing the COVID-19 dataset}
\begin{figure}[H]
  \includegraphics[width=.5\linewidth]{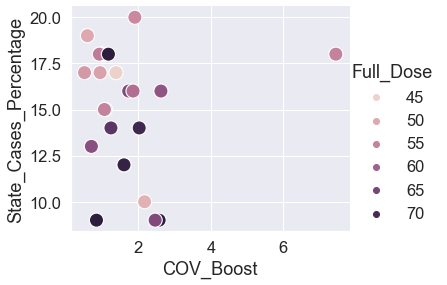}
  \caption{The percentage of each state's population who obtained full- dose and booster shots}
  \label{fig:10}
\end{figure}

\begin{figure}[H]
  \includegraphics[width=.9\linewidth]{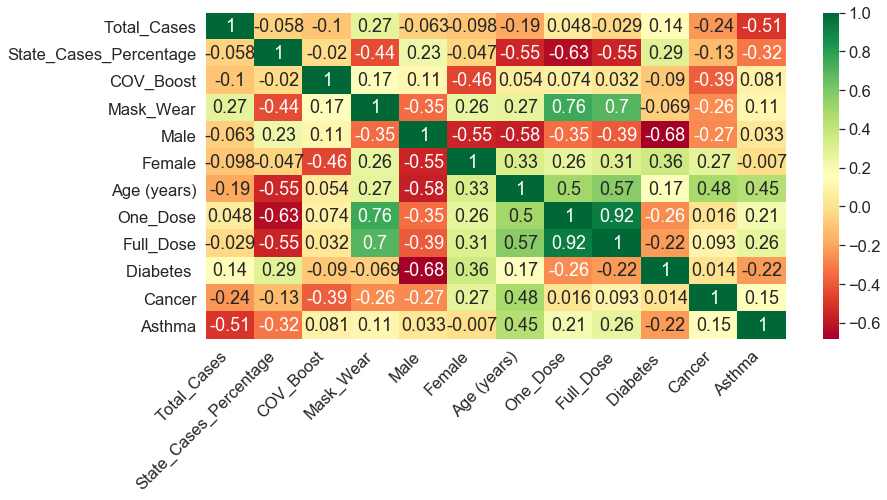}
  \caption{The percentage of each state's population who obtained full- dose and booster shots}
  \label{fig:heatmap}
\end{figure}

 To finish the analysis, this paper considers a final dataset for the purpose of diminishing any doubt in the efficacy of the COVID-19 vaccines and boosters for protecting patients against the dangers associated from COVID-19 and any other underlying health conditions that could strengthen the symptoms. Figure \ref{fig:10} provides a scatterplot relating COV-Boost (the percentage of the state population who received a COVID-19 booster shot) and Full-Dose (the percentage of the state population who received both doses of either the Pfizer or Moderna COVID-19 vaccine shots) with the percentage of the state population with COVID-19 cases. Evident by the distribution, there appears to be a negative correlation between COV-Boost and State-Cases-Percentage, indicating that as the percentage of the state population who received a COVID-19 booster shot increases the percentage of the state population with COVID-19 cases decreases.
Additionally, although states with higher percentages of full dose vaccination among the population tend to have lower percentages of people who have received the COVID-19 booster shot, the total COVID-19 case percentage still appears to be low. For example, states with 70\% or greater full dose vaccination rates tend to have low COVID-19 case percentages among the population, specifically below 14\%. Therefore, it is plausible to presume that the variables Full-Dose and COV-Boost have a negative correlation with the percentage of the state population who have COVID-19. As such, this paper provides additional evidence of this negative correlation by inducing a correlation matrix heat map of each variable considered in this dataset (see Figure \ref{fig:heatmap}). Concerning the correlation coefficients of COV-Boost and Full-Dose in relation to State-Cases-Percentage, both values are negative, indicating that as the percentages of the population who received the COVID-19 booster or both doses of the COVID-19 vaccine increase, the percentage of the population who have contracted COVID-19 will decrease.  Additionally, this conclusion can be confidently drawn since the correlation coefficients for One-Dose and Full-Dose are relatively close to -1, indicating that these effects have a strong, negative relationship with the percentage of the population with COVID-19. 

\begin{figure}[H]
\centering
\begin{subfigure}{.5\textwidth}
  \centering
    \caption{The relationship between common underlying health conditions and the percentage of the population with COVID-19}
  \includegraphics[width=\linewidth]{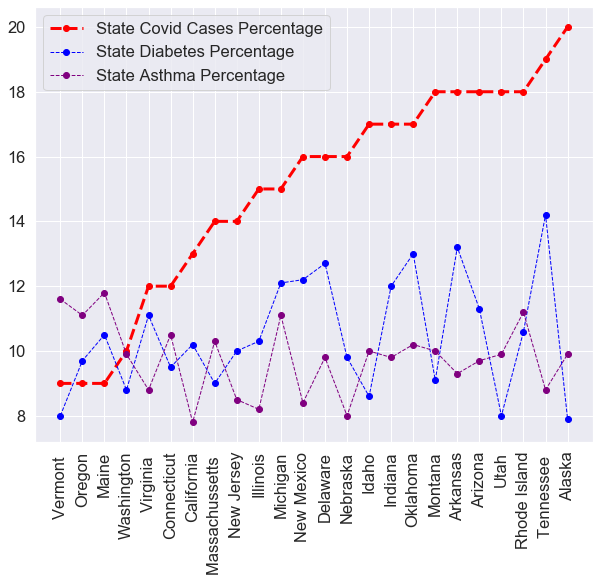}
  \label{fig:cat10w}
\end{subfigure}%
\begin{subfigure}{.5\textwidth}
  \centering
    \caption{The relationship between percentage of each state’s population who wear a mask, the percentage age of the population that have both doses of the COVID-19 vaccine, and the state's median age}
  \includegraphics[width=.92\linewidth]{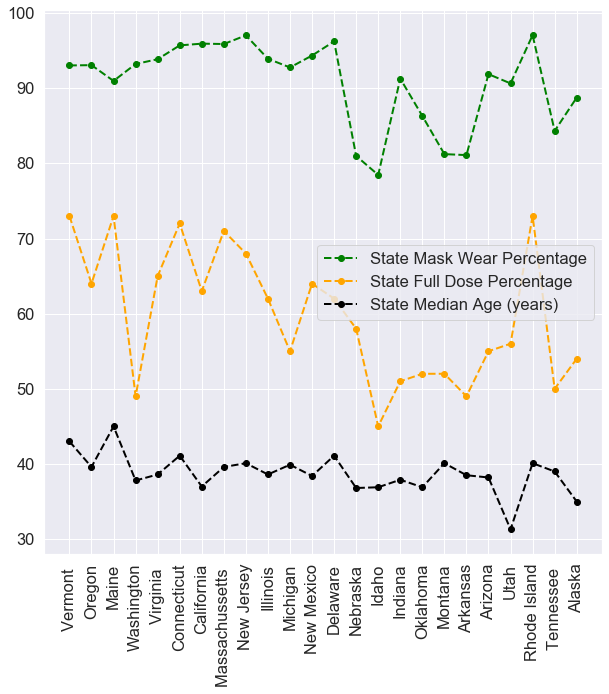}
  \label{fig:cat100w}
\end{subfigure}
\caption{Line graphs}
\label{fig:linegraph}
\end{figure}

Apart from the effects representing the percentages of the population having one dose, two doses, or a booster shot of the COVID-19 vaccines demonstrating a strong, negative relationship with the percentage of the population having COVID-19, average age (“Age”) of the state population also has a strong, negative correlation with the variable State-Cases-Percentage. As such, the correlation value -0.55 indicates that as the average age of the population increases in the specified state, the percentage of the total cases among the state population will tend to decrease. This negative correlation could possibly be attributed to the older parts of the population being more cautious about the severity of the symptoms associated with COVID-19, or simply put, younger subgroups of the populations (i.e., two to five years of age) are not eligible to receive the Pfizer or Moderna COVID-19 vaccines. 
Because it is likely that underlying health conditions tend to be prominent in older subgroups of the population, Figure \ref{fig:linegraph} also displays two line graphs: a line graph investigating the relationship between common underlying health conditions and the percentage of the population with COVID-19, and an additional line graph relating the percentage of each state’s population who wear a mask, the percentage age of the population that have both doses of the COVID-19 vaccine, and the state's median age. Evident from the results of the second line graph. From the first graph, it is evident that states with a higher percentage of the population with diabetes also tends to have a higher percentage of the population with COVID-19, confirming the reoccurring relationship between diabetes mellitus and COVID-19. Evident from the results of the second line graph, it appears that states with greater percentages of people who wear masks also tend to have high percentages of people who are fully vaccinated. This result makes statistical sense because the correlation coefficient relating the percentage of the state’s population who wears a mask to the percentage of the population who has COVID-19 is negative. Applicably, as the percentage of the state’s population who wears a mask increase, the percentage of the population who has COVID-19 will decrease.
In all, from the results of this dataset, it is evident that states with higher proportions of people who are fully vaccinated, obtained an additional COVID-19 booster, and wear masks, tend to have lower proportions of people who have COVID-19.

\section{Discussion}\label{section4}
The results of this study have various applications to the world we now live in because of the COVID-19 pandemic. First, the results from the first three datasets suggested that there is a strong positive relationship between specific underlying health conditions and the COVID-19 total death percentage. For example, it is strongly apparent that there is a significant, positive relationship between the diabetes mellitus and COVID-19. Additionally, it is also concluded that malnutrition, such as obesity and tobacco use, also contribute to the COVID-19 total death count percentage in the United States and Mexico.

Considering the serenity of the positive relationship between underlying health conditions and illnesses with COVID-19, this paper also demonstrate the efficacy of the COVID-19 vaccines and boosters as preventive measures against the symptoms associated with COVID-19. More specifically, the analysis provided by the “adjusted” dataset suggests paramount conclusions about how to reduce the total COVID-19 case and death count. In summary, it was concluded that the states with higher proportions of people who are fully vaccinated, obtained an additional COVID-19 booster, and wear masks, tend to have lower proportions of people who have COVID-19. Therefore, it is profoundly evident that obtaining full COVID-19 vaccine dosages, receiving a booster shot, and wearing a mask will tend to reduce the likelihood of contracting COVID-19.

Considering these results, this paper makes two important suggestions about the hypotheses proposed in the introduction. First, with regards to the question, “For the typical American citizen, does an underlying health condition contribute to the fatality rate of COVID-19 across the United States,” this paper demonstrated that there is a profound and positive relationship between several, relevant underlying health conditions and the severity of the symptoms associated with COVID-19. In particular, individuals with Diabetes Mellitus will tend to have more severe, life-threatening symptoms associated with COVID-19 in relation to individuals who do not have this underlying health condition. To further support this broad conclusion concerning the relationship between selected underlying health conditions and COVID-19, this paper also demonstrates similar conclusions using patient information in Mexico. More specifically, underlying conditions such as diabetes, asthma, cardiovascular, hypertension, renal-chronic and tobacco related diseases were found to have a profound relationship with COVID-19 related deaths. Therefore, these results indicate that regardless of the ethnicity and
country, there is a strong relationship between underlying health conditions and the total COVID-19 death count.

As a segue to the second hypothesis considered, this paper sought to connect the apparent relationship between selected underlying health conditions and COVID-19 with the efficacy of the COVID-19 vaccine and booster shots. In particular, this paper serves as a plea for individuals to receive the COVID-19 vaccines and booster shots as preventive measures for possible hospitalization as well as to mitigate the severity of the symptoms associated with COVID-19 for
individuals with underlying health conditions. According to the results presented by this study, the COVID-19 vaccines and booster shots do in fact tend to provide sufficient protection against the array of COVID-19 variants for the typical American. In short, there a strong, negative relationship between the COVID-19 vaccination rates and the COVID-19 confirmed cases. As such, as the vaccination rates increase across the nation, the COVID-19 total case count will tend to decrease. Therefore, individuals should carefully consider the relationship between the apparent contagiousness associated with COVID-19 and the number of people who have received a COVID-19 vaccine and booster shot.

In all, this paper seeks to extend its conclusions by proposing that individuals who are unvaccinated in the current state of the pandemic will be at a higher risk of contracting COVID-19. Furthermore, individuals who are unvaccinated and have an underlying health condition could experience more severe and fatal symptoms associated with COVID-19, ultimately increasing the rate of hospitalizations. Therefore, regarding the fundamental duty and respectfulness expected of an American citizen, receiving the COVID-19 vaccine and booster shot will only aid in ending this pandemic rather than prolonging it.

\section{Conclusion}\label{section5}
In this paper, we have utilized several statistical tests and machine learning methods to analyze the relationship between some underlying medical conditions and COVID-19 susceptibility.  Our findings suggest that there is a significant, positive relationship between the diabetes mellitus and COVID-19. Moreover, it was also concluded that malnutrition, such as obesity and tobacco use, also contribute to the COVID-19 total death count percentage in the United States and Mexico. Additionally, we have discussed the efficacy of the COVID-19 vaccines and boosters as preventive measures against the symptoms associated with COVID-19. The paper also concludes that the states with higher proportions of people who are fully vaccinated, obtained an additional COVID-19 booster, and wear masks, tend to have lower proportions of people who have COVID-19. Therefore, it is profoundly evident that obtaining full COVID-19 vaccine dosages, receiving a booster shot, and wearing a mask will tend to reduce the likelihood of contracting COVID-19.

\bibliography{mybibfile}

\begin{thebibliography}{10}
\expandafter\ifx\csname url\endcsname\relax
  \def\url#1{\texttt{#1}}\fi
\expandafter\ifx\csname urlprefix\endcsname\relax\def\urlprefix{URL }\fi
\expandafter\ifx\csname href\endcsname\relax
  \def\href#1#2{#2} \def\path#1{#1}\fi
\bibitem{CDC 1}
“Comparative Effectiveness of Moderna, Pfizer-Biontech, and Janssen (Johnson \& Johnson) Vaccines in Preventing COVID-19 Hospitalizations among Adults without Immunocompromising Conditions - United States, March–August 2021.” Centers for Disease Control and Prevention, Centers for Disease Control and Prevention, 23 Sept. 2021, \url{www.cdc.gov/mmwr/volumes/70/wr/mm7038e1.htm}.
\bibitem{1}
Ruder, Sebastian. "An overview of gradient descent optimization algorithms." arXiv preprint arXiv:1609.04747 (2016).
\bibitem{2}
Bottou, Léon. "Stochastic gradient descent tricks." Neural networks: Tricks of the trade. Springer, Berlin, Heidelberg, 2012. 421-436.
\bibitem{3}
Gill, Philip E., Walter Murray, and Margaret H. Wright. Practical optimization. Society for Industrial and Applied Mathematics, 2019.
\bibitem{4}
Nocedal, Jorge, and Stephen Wright. Numerical optimization. Springer Science \& Business Media, 2006.
\bibitem{5}
Conn, Andrew R., Nicholas IM Gould, and Philippe L. Toint. Trust region methods. Society for Industrial and Applied Mathematics, 2000.
\bibitem{6}
Brust, Johannes, Jennifer B. Erway, and Roummel F. Marcia. "On solving L-SR1 trust-region subproblems." Computational Optimization and Applications 66.2 (2017): 245-266.
\bibitem{7}
Erway, Jennifer B., and Roummel F. Marcia. "Limited-memory BFGS systems with diagonal updates." Linear algebra and its applications 437.1 (2012): 333-344.
\bibitem{8}
Rezapour, Mostafa. Trust-Region Methods for Unconstrained Optimization Problems. Washington State University, 2020.
\bibitem{9}
Rezapour, Mostafa, and Thomas J. Asaki. "Adaptive trust-region algorithms for unconstrained optimization." Optimization Methods and Software (2019): 1-23.
\bibitem{10}
Erway, Jennifer B., and Mostafa Rezapour. "A New Multipoint Symmetric Secant Method with a Dense Initial Matrix." arXiv preprint arXiv:2107.06321 (2021).

\bibitem{11-n}
\url{https://www.kaggle.com/kaggle-survey-2020}
\bibitem{11-n1}
Natekin, Alexey, and Alois Knoll. "Gradient boosting machines, a tutorial." Frontiers in neurorobotics 7 (2013): 21.

\bibitem{34}
Chen, Tianqi, and Carlos Guestrin. "Xgboost: A scalable tree boosting system." Proceedings of the 22nd acm sigkdd international conference on knowledge discovery and data mining. 2016


\bibitem{35}
Chen, Tianqi, et al. "Xgboost: extreme gradient boosting." R package version 0.4-2 1.4 (2015): 1-4.

\bibitem{36}
Ke, Guolin, et al. "Lightgbm: A highly efficient gradient boosting decision tree." Advances in neural information processing systems 30 (2017): 3146-3154.
\bibitem{37}
Dorogush, Anna Veronika, Vasily Ershov, and Andrey Gulin. "CatBoost: gradient boosting with categorical features support." arXiv preprint arXiv:1810.11363 (2018).
\bibitem{37-1}
Prokhorenkova, Liudmila, et al. "CatBoost: unbiased boosting with categorical features." Advances in neural information processing systems 31 (2018).
\bibitem{friedman2001greedy}
Friedman, Jerome H. "Greedy function approximation: a gradient boosting machine." Annals of statistics (2001): 1189-1232.
\bibitem{breiman1996bagging}
Breiman, Leo. "Bagging predictors." Machine learning 24.2 (1996): 123-140.
\bibitem{GBDTTREE1}
Manish Mehta, Rakesh Agrawal, and Jorma Rissanen. Sliq: A fast scalable classifier for data mining. In International Conference on Extending Database Technology, pages 18–32. Springer, 1996.
\bibitem{GBDTTREE2}
John Shafer, Rakesh Agrawal, and Manish Mehta. Sprint: A scalable parallel classi er for data mining. In Proc. 1996 Int. Conf. Very Large Data Bases, pages 544–555. Citeseer, 1996.
\bibitem{GBDTTREE3}
Sanjay Ranka and V Singh. Clouds: A decision tree classifier for large datasets. In Proceedings of the 4th Knowledge Discovery and Data Mining Conference, pages 2–8, 1998.
\bibitem{GBDTTREE4}
Ruoming Jin and Gagan Agrawal. Communication and memory efficient parallel decision tree construction. In Proceedings of the 2003 SIAM International Conference on Data Mining, pages 119–129. SIAM, 2003.
\bibitem{GBDTTREE5}
Ping Li, Christopher JC Burges, Qiang Wu, JC Platt, D Koller, Y Singer, and S Roweis. Mcrank: Learning to rank using multiple classification and gradient boosting. In NIPS, volume 7, pages 845–852, 2007.
\bibitem{shi2007best}
Shi, Haijian. Best-first decision tree learning. Diss. The University of Waikato, 2007.

\bibitem{Breiman}
Breiman, Leo. "Random forests." Machine learning 45.1 (2001): 5-32.

\bibitem{laurent1976constructing}
Laurent, Hyafil, and Ronald L. Rivest. "Constructing optimal binary decision trees is NP-complete." Information processing letters 5.1 (1976): 15-17.
\bibitem{11}
Kushwaha, Shashi, et al. "Significant applications of machine learning for COVID-19 pandemic." Journal of Industrial Integration and Management 5.04 (2020): 453-479.

\bibitem{12}
De Felice, Francesca, and Antonella Polimeni. "Coronavirus disease (COVID-19): a machine learning bibliometric analysis." in vivo 34.3 suppl (2020): 1613-1617.
\bibitem{13}
Alimadadi, Ahmad, et al. "Artificial intelligence and machine learning to fight COVID-19." Physiological genomics 52.4 (2020): 200-202.
\bibitem{14}
Elaziz, Mohamed Abd, et al. "New machine learning method for image-based diagnosis of COVID-19." Plos one 15.6 (2020): e0235187.
\bibitem{15}
Punn, Narinder Singh, Sanjay Kumar Sonbhadra, and Sonali Agarwal. "COVID-19 epidemic analysis using machine learning and deep learning algorithms." MedRxiv (2020).
\bibitem{16}
Sujath, R., Jyotir Moy Chatterjee, and Aboul Ella Hassanien. "A machine learning forecasting model for COVID-19 pandemic in India." Stochastic Environmental Research and Risk Assessment 34 (2020): 959-972.
\bibitem{17}
Lalmuanawma, Samuel, Jamal Hussain, and Lalrinfela Chhakchhuak. "Applications of machine learning and artificial intelligence for Covid-19 (SARS-CoV-2) pandemic: A review." Chaos, Solitons \& Fractals 139 (2020): 110059.
\bibitem{18}
Cheng, Fu-Yuan, et al. "Using machine learning to predict ICU transfer in hospitalized COVID-19 patients." Journal of clinical medicine 9.6 (2020): 1668.
\bibitem{19}
Rezapour, Mostafa, and Lucas Hansen. "A Machine Learning Analysis of COVID-19 Mental Health Data." arXiv preprint arXiv:2112.00227 (2021).
\bibitem{20}
Rezapour, Mostafa. "Hidden Effects of COVID-19 on Healthcare Workers: A Machine Learning Analysis." arXiv preprint arXiv:2112.06261 (2021).
\bibitem{21}
Ah Monthly Provisional Counts of Deaths by Age Group and HHS Region for Select Causes of Death, 2019-2021 \url{https://data.cdc.gov/NCHS/AH-Monthly-Provisional-Counts-of-Deaths-by-Age-Gro/ezfr-g6hf}

\bibitem{22}
 “Alabama Health Center Covid-19 Survey Summary Report.” Bureau of Primary Health Care, 8 Apr. 2020, \url{bphc.hrsa.gov/emergency-response/coronavirus-health-center-data/al}.
\bibitem{23}
 Cancer Rates by State 2021, \url{worldpopulationreview.com/state-rankings/cancer-rates-by-state}.
\bibitem{24} 
“Comparative Effectiveness of Moderna, Pfizer-Biontech, and Janssen (Johnson \& Johnson) Vaccines in Preventing COVID-19 Hospitalizations among Adults without Immunocompromising Conditions - United States, March–August 2021.” Centers for Disease Control and Prevention, Centers for Disease Control and Prevention, 23 Sept. 2021, \url{www.cdc.gov/mmwr/volumes/70/wr/mm7038e1.htm}.
\bibitem{25} 
“Diabetes in the United States.” The State of Childhood Obesity, \url{stateofchildhoodobesity.org/diabetes/}.
\item Elflein, John. “U.S. COVID-19 Case Rate by State.” Statista, 16 Dec. 2021, \url{www.statista.com/statistics/1109004/coronavirus-covid19-cases-rate-us-americans-by-state/}.
\bibitem{26} 
Health Center Covid-19 Operations, \url{data.hrsa.gov/topics/health-centers/covid-operational-capacity}.
\bibitem{27}
 “List of U.S. States and Territories by Median Age.” Wikipedia, Wikimedia Foundation, 11 Apr. 2021, \url{en.wikipedia.org/wiki/List_of_U.S._states_and_territories_by_median_age}.
\bibitem{28}
 “Most Recent Asthma State Data.” Centers for Disease Control and Prevention, Centers for Disease Control and Prevention, 30 Mar. 2021, \url{www.cdc.gov/asthma/most_recent_data_states.htm}.
\bibitem{29} 
“Population Distribution by Sex.” KFF, 23 Oct. 2020, \url{https://www.kff.org/other/state-indicator/distribution-by-sex/?currentTimeframe=0&sortModel=\%7B\%22colId\%22:\%22Location\%22,\%22sort\%22:\%22asc\%22\%7D}.
\bibitem{30}
 “U.S. Coronavirus Map: Tracking the Trends.” Mayo Clinic, Mayo Foundation for Medical Education and Research, \url{www.mayoclinic.org/coronavirus-covid-19/map}.

\bibitem{31}
“Ah Provisional Covid-19 Deaths by Educational Attainment, Race, Sex, and Age.” Centers for Disease Control and Prevention, Centers for Disease Control and Prevention, \url{data.cdc.gov/NCHS/AH-Provisional-COVID-19-Deaths-by-Educational-Atta/3ts8-hsrw}. 
\bibitem{31e}
Ah Monthly Provisional Counts of Deaths by Age Group and HHS Region for Select Causes of Death, 2019-2021.” Centers for Disease Control and Prevention, Centers for Disease Control and Prevention, data.cdc.gov/NCHS/AH-Monthly-Provisional-Counts-of-Deaths-by-Age-Gro/ezfr-g6hf.
\bibitem{31ee}
“Comparative Effectiveness of Moderna, Pfizer-Biontech, and Janssen (Johnson \& Johnson) Vaccines in Preventing COVID-19 Hospitalizations among Adults without Immunocompromising Conditions - United States, March–August 2021.” Centers for Disease Control and Prevention, Centers for Disease Control and Prevention, 23 Sept. 2021, \url{www.cdc.gov/mmwr/volumes/70/wr/mm7038e1.htm}.
\bibitem{32}
Mukherjee, Tanmoy. “Covid-19 Patient Pre-Condition Dataset.” Kaggle, 22 July 2020, \url{www.kaggle.com/tanmoyx/covid19-patient-precondition-dataset}. 
\bibitem{33}
\url{https://github.com/MostafaRezapour/A-machine-learning-analysis-of-the-relationship-between-some-underlying-medical-conditions-and-COVID}.


\bibitem{38}
Al Daoud, Essam. "Comparison between XGBoost, LightGBM and CatBoost using a home credit dataset." International Journal of Computer and Information Engineering 13.1 (2019): 6-10.


\end{thebibliography}

\end{document}